\documentclass[mathpazo]{cicp}

\usepackage{amsmath,amsfonts,amsthm,amssymb}
\usepackage{algorithm2e}
\usepackage{bm,dsfont}
\usepackage{booktabs}
\usepackage{graphicx}
\usepackage{array}
\usepackage{xcolor}
\usepackage{sidecap}
\usepackage{listings}
\usepackage{times}
\usepackage{url}
\usepackage{hyperref}
\graphicspath{{figs/}}
\newtheorem{corollary}[theorem]{Corollary}

\newcommand{\N}{\mathcal{N}}
\newcommand{\D}{\mathcal{D}}

\newcommand{\R}{\mathbb{R}}

\newcommand{\x}{\textbf{x}}
\newcommand{\W}{\bm{W}}
\DeclareMathOperator*{\argmin}{argmin}

\definecolor{codegreen}{rgb}{0,0.6,0}
\definecolor{codegray}{rgb}{0.5,0.5,0.5}
\definecolor{codepurple}{rgb}{0.58,0,0.82}
\definecolor{backcolour}{rgb}{0.95,0.95,0.92}

\lstdefinestyle{mystyle}{
    backgroundcolor=\color{backcolour},   
    commentstyle=\color{codegreen},
    keywordstyle=\color{magenta},
    numberstyle=\tiny\color{codegray},
    stringstyle=\color{codepurple},
    basicstyle=\ttfamily\scriptsize,
    breakatwhitespace=false,         
    breaklines=true,                 
    captionpos=b,                    
    keepspaces=true,                 
    language=Python,
    numbers=left,                    
    numbersep=5pt,                  
    showspaces=false,                
    showstringspaces=false,
    showtabs=false,                  
    tabsize=2
}
\begin{document}

\title{Dying ReLU and Initialization: Theory and Numerical Examples}

\author[Lu L et.~al.]{Lu Lu\affil{1},
       Yeonjong Shin\affil{2}\comma\corrauth, Yanhui Su\affil{3}, and George Em Karniadakis\affil{2}}
\address{\affilnum{1}\ Department of Mathematics, Massachusetts Institute of Technology, Cambridge, MA 02139, USA \\
\affilnum{2}\ Division of Applied Mathematics, Brown University, Providence, RI 02912, USA \\
         \affilnum{3}\ College of Mathematics and Computer Science, Fuzhou University, Fuzhou, Fujian 350116, China}
\emails{{\tt lu\_lu@mit.edu} (L.~Lu), {\tt yeonjong\_shin@brown.edu} (Y.~Shin), {\tt suyh@fzu.edu.cn} (Y.~Su),
        {\tt george\_karniadakis@brown.edu} (G.~E.~Karniadakis). L. Lu and Y. Shin contributed equally to this work.}

\begin{abstract}
The dying ReLU refers to the problem when ReLU neurons become inactive and only output 0 for any input.
There are many empirical and heuristic explanations
of why ReLU neurons die. However,
little is known about its theoretical analysis.
In this paper, we rigorously prove that a deep ReLU network will eventually die in probability as the depth goes to infinite.
Several methods have been proposed to alleviate the dying ReLU. Perhaps, one of the simplest treatments is to modify the initialization procedure. 
One common way of initializing weights and biases uses symmetric probability distributions, which suffers from the dying ReLU.
We thus propose a new initialization procedure, namely, a randomized asymmetric initialization.
We show that the new initialization can effectively prevent the dying ReLU. All parameters required for the new initialization are theoretically designed.
Numerical examples are provided to demonstrate the effectiveness of the new initialization procedure.
\end{abstract}



\maketitle

\section{Introduction} \label{sec:intro}
The rectified linear unit (ReLU), $\max\{x,0\}$,
is one of the most successful and widely-used activation functions in deep learning
\cite{lecun2015deep, ramachandran2017searchingAct, nair2010rectified}.
The success of ReLU is based on its superior training performance \cite{glorot2011deep, sun2015deeply} over other activation functions such as the logistic sigmoid and the hyperbolic tangent \cite{glorot2010understanding, lecun1998efficient}.
The ReLU has been used in various applications including image classification \cite{krizhevsky2012imagenet, szegedy2015going}, natural language processes \cite{maas2013rectifier},  speech recognition \cite{hinton2012deep}, 
and game intelligence \cite{silver2016mastering}, 
to name a few.

The use of gradient-based optimization is inevitable in training deep neural networks.
It has been widely known that 
the deeper a neural network is, the harder it is to train
\cite{srivastava2015training, du2018gradient}.
A fundamental difficulty in training deep neural networks is the vanishing and exploding gradient problem \cite{poole2016exponential, hanin2018whichNN, chen2018dynamical}.
The dying ReLU is a kind of vanishing gradient, which refers to a problem when ReLU neurons become inactive and only output 0 for any input.
It has been known as one of the obstacles in training deep feed-forward ReLU neural networks \cite{trottier2017parametric, agarap2018deep}.
To overcome this problem, a number of methods have been proposed.
Broadly speaking, these can be categorized into three general approaches.
One approach modifies the network architectures.
This includes but not limited to the changes in the number of layers, the number of neurons, network connections, and activation functions.
In particular, many activation functions have been proposed to replace the ReLU \cite{maas2013rectifier,he2015delving,clevert2015fast, klambauer2017self}.
However, the performance of other activation functions  varies on different tasks and data sets \cite{ramachandran2017searchingAct} 
and it typically requires a parameter to be turned.
Thus, the ReLU remains one of the popular activation functions due to its simplicity and reliability. 
Another approach introduces additional training steps.
This includes several normalization techniques \cite{ioffe2015batch, salimans2016weight, ulyanov2016instance, ba2016layer, wu2018group} and dropout \cite{srivastava2014dropout}.
One of the most successful normalization techniques is 
the batch normalization \cite{ioffe2015batch}.
It is a technique that inserts layers into the deep neural network 
that transform the output for the batch to be zero mean unit variance.
However, batch normalization increases by 30\% the computational overhead to each iteration \cite{mishkin2015all}.
The third approach modifies only weights and biases initialization procedure without changing any network architectures or introducing additional training steps
\cite{lecun1998efficient, glorot2010understanding, he2015delving, saxe2013exact, mishkin2015all}.
The third approach is the topic of our work presented in this paper.

The intriguing ability of gradient-based optimization
is perhaps one of the major contributors to the success of deep learning.
Training deep neural networks using gradient-based optimization
fall into the noncovex nonsmooth optimization.
Since a gradient-based method is either a first- or a second-order method, and once converged,
the optimizer is either a local minimum or a saddle point.
The authors of \cite{fukumizu2000local} proved that the existence of local minima poses a serious problem in training neural networks. 
Many researchers have been putting immense efforts to 
mathematically understand the gradient method and its ability to solve
nonconvex nonsmooth problems.
Under various assumptions, especially on the landscape,
many results claim that the gradient method can find a global minimum, can escape saddle points, and can avoid spurious local minima \cite{lee2016gradient, amari2006singularities, ge2015escaping, ge2016matrix, zhou2017critical, wu2018no, yun2018small, du2017gradexp, du2017gradonehidden, du2018gradient, jin2017escape}.
However, these assumptions do not always hold and are provably false for deep neural networks \cite{safran2018spurious, kawaguchi2016deep, arora2018convergence}.
This further limits our understanding on what contributes to the success of the deep neural networks.
Often, theoretical conditions are impossible to be met in practice.

\begin{figure}[htbp]
\centering
\includegraphics{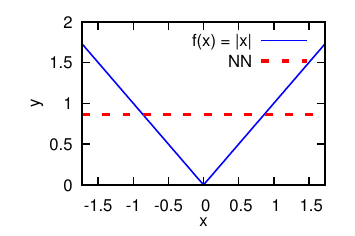}
\caption{An approximation result for $f(x)=|x|$ using a 10-layer ReLU neural network of width 2. Among 1,000 independent simulations, this trained result is obtained with more than 90\% probability. One of the most popular initialization procedures \cite{he2015delving} is employed.}
\label{fig:absx-intro}
\end{figure}

Where to start the optimization process plays a critical role in training
and has a significant effect on the trained result \cite{nesterov2013introductory}.
This paper focuses on a particular kind of bad local minima
due to a bad initialization.
Such a bad local minimum causes the dying ReLU.
Specifically, we consider the worst case of dying ReLU, where 
the entire network dies, i.e., the network becomes a constant function.
We refer this as \textit{the dying ReLU neural networks} (NNs).
This phenomenon could be well illustrated by a simple example.
Suppose $f(x) = |x|$ is a target function we want to approximate using a ReLU network. Since $|x| = \text{ReLU}(x) + \text{ReLU}(-x)$, a 2-layer ReLU network of width 2 can exactly represent $|x|$. However, when we train a deep ReLU network, we frequently observe that the network is collapsed. This trained result is shown in Fig.~\ref{fig:absx-intro}.
Our 1,000 independent simulations show that there is a high probability (more than 90\%) for the deep ReLU network to collapse to a constant function. In this example, we employ a 10-layer ReLU network of width 2 which should perfectly recover $f(x)=|x|$.

Almost all common initialization schemes in training deep neural networks
use symmetric probability distributions around 0.
For example, zero mean uniform distributions and zero mean normal distributions were proposed and used in \cite{lecun1998efficient, glorot2010understanding, he2015delving}.
We show that when weights and biases are initialized from symmetric probability distributions around 0,
the dying ReLU NNs occurs in probability as the number of depth goes to infinite.
To the best of our knowledge, this is the first theoretical work on the dying ReLU.
This result partially explains why training \textit{extremely} deep networks is challenging.
Furthermore, it says that the dying ReLU is inevitable as long as the network is deep enough.
Also, our result implies that it is the network architecture 
that decides whether an initialization procedure is good or bad.
Our analysis reveals that a specific network architecture
can avoid the dying ReLU NNs with high probability.
That is, for any $\delta > 0$, 
when a symmetric initialization is used
and $L = \Omega(\log_2 (N/\delta))$ is satisfied 
where $L$ is the number of depth and $N$ is the number of width at each layer,
with probability $1-\delta$, 
the dying ReLU NNs will not happen.
This explains why wide networks are popularly used in practice.
However, the use of deep and wide networks 
requires one to train a gigantic number of 
parameters and this could be computationally 
very expensive and time consuming.

On the other hands, deep and narrow networks 
play fundamental roles both in theory 
and in practice.
For example, in the work of \cite{he2020relu}, it was shown that 
deep and narrow ReLU networks are inevitable in constructing finite element's basis functions.
This represents an application of deep ReLU networks in finite element method for solving PDEs.
Also, many theoretical works \cite{petersen2017optimal,yarotsky2017error,hanin2017approximating} on the expressivity of ReLU networks rely heavily on deep and narrow networks 
in approximating polynomials through sparse concatenations.
Despite the importance of 
deep and narrow networks,
due to aforementioned reasons, 
they have not been successfully employed in practice.
Thereby, we propose an initialization scheme, 
namely, 
a randomized asymmetric initialization (RAI),
to directly overcome the dying ReLU.
On the theoretical side, 
we show that RAI has a smaller upper bound of the probability of the dying ReLU NNs
than standard symmetric initialization
such as the He initialization \cite{he2015delving}.
All hyper-parameters for RAI are theoretically driven 
to avoid the exploding gradient problem. 
This is done by the second moment analysis \cite{he2015delving, hanin2018whichNN, poole2016exponential}.


The rest of the paper is organized as follows.
After setting up notation and terminology in Section~\ref{sec:set-up},
we present the main theoretical results in Section~\ref{sec:theory}.
In Section~\ref{sec:new-init}, upon introducing a randomized asymmetric initialization,
we discuss its theoretical properties.
Numerical examples are provided in Section~\ref{sec:example}, before the conclusion in
Section~\ref{sec:conclusion}.

\section{Mathematical setup} \label{sec:set-up}

Let $\N^L: \mathbb{R}^{d_{\text{in}}} \to \mathbb{R}^{d_{\text{out}}}$ 
be a feed-forward neural network
with $L$ layers and $N_\ell$ neurons in the $\ell$-th layer ($N_0 = d_{\text{in}}$, $N_L = d_{\text{out}}$). 
Let us denote the weight matrix and bias vector in the $\ell$-th layer
by $\bm{W}^\ell \in \mathbf{R}^{N_\ell \times N_{\ell-1}}$ and  $\mathbf{b}^\ell \in \mathbb{R}^{N_\ell}$, respectively. 
Given an activation function $\phi$ which is applied element-wise,
the feed-forward neural network is recursively defined as follows: $\mathcal{N}^1(\textbf{x}) = \bm{W}^{1}\textbf{x} + \bm{b}^{1}$ and
\begin{align}
	\mathcal{N}^{\ell}(\textbf{x}) &= \bm{W}^{\ell}\phi(\mathcal{N}^{\ell-1}(\textbf{x})) + \bm{b}^{\ell}
	\in \mathbb{R}^{N_\ell}, 
	\qquad \text{for} \quad 2 \le \ell \le L.
\end{align}
Here $\N^L$ is called a $L$-layer neural network or a $(L-1)$-hidden layer neural network.
In this paper, the rectified linear unit (ReLU) activation function is employed, i.e.,
$$
\phi(\x) = \text{ReLU}(\x) :=\left(\max\{x_1, 0\}, \cdots, \max\{x_{\text{fan-in}}, 0\} \right),
\quad
\text{where} \quad \x = (x_1,\cdots,x_{\text{fan-in}}).
$$
Let $\vec{\bm{N}} = (N_0,N_1,\cdots, N_L)$ be a vector representing 
the network architecture.
Let $\bm{\theta}_{\vec{\bm{N}}} = \{\bm{W}^\ell, \bm{b}^\ell\}_{1 \le \ell \le L}$ be the set of all weight matrices and bias vectors.
Let $\mathcal{T}_m = \{\x_i, y_i\}_{1\le i \le m}$
be the set of $m$ training data
and let $\mathcal{D} = \{\x_i\}_{i=1}^m$ be the training input data.
We assume that $\mathcal{D} \subset B_r(0)$ for some $r > 0$.
Given $\mathcal{T}_m$, in order to train $\bm{\theta}_{\vec{\bm{N}}}$, we consider 
the standard loss function $\mathcal{L}(\bm{\theta}_{\vec{\bm{N}}}, \mathcal{T}_m)$:
\begin{equation} \label{def:loss-fns}
    \mathcal{L}(\bm{\theta}_{\vec{\bm{N}}}, \mathcal{T}_m)
    = \frac{1}{m}\sum_{(\x,y) \in \mathcal{T}_m} \ell(\N^L(\x; \bm{\theta}_{\vec{\bm{N}}}),y),
\end{equation}
where $\ell:\mathbb{R}^{d_{\text{out}}} \times \mathbb{R}^{d_{\text{out}}} \to \mathbb{R}$
is a loss criterion.
In training neural networks, the gradient-based optimization is typically employed to minimize the loss 
$\mathcal{L}$.
The first step for training would be to initialize weight matrices and bias vectors.
Typically, they are initialized according to certain probability distributions. For example, uniform distributions around 0 or zero-mean normal distributions are common choices.

In this paper, 
we focus on the worst case of dying ReLU, where 
the entire network dies, i.e., the network becomes a constant function.
We refer this as the dying ReLU neural network.
We then define two phases:
(1) a network is dead before training,
and
(2) a network is dead after training.
The phase 1 implies the phase 2, but not vice versa.
When the phase 1 happens, we say \textit{the network is born dead} (BD).

\section{Dying probability bounds and asymptotic properties} \label{sec:theory}

We investigate the probability of the dying ReLU neural networks at random initialization.
Let $(\Omega,\mathcal{F}, P)$ be a probability space.
For a network architecture $\vec{\bm{N}}_L = (d_{\text{in}}, N_1, \cdots,N_{L-1}, d_{\text{out}})$,
let $\bm{\theta}_{\vec{\bm{N}}_L}:\Omega \to \mathbb{R}^{|\vec{\bm{N}}_L|}$
be a vector-valued random variable,
where $|\vec{\bm{N}}_L| := \sum_{\ell = 1}^L N_{\ell}(N_{\ell-1}+1)$ is the total number of network parameters.
Let us consider the following BD event:
\begin{equation} \label{def:dyingNN-event}
    \mathfrak{J}_{\vec{\bm{N}}_L} :=
    \{ \omega \in \Omega \hspace{0.1cm} | \hspace{0.1cm} 
    \mathcal{N}^L(\x; \bm{\theta}_{\vec{\bm{N}}_L}(\omega))
    \text{ is a constant function in } \mathcal{D}
    \}.
\end{equation}
Then, $P(\mathfrak{J}_{\vec{\bm{N}}_L})$
is the BD probability (BDP).
The following theorem shows that a deep ReLU network will eventually be BD in probability as the number of depth $L$ goes to infinity.


\begin{theorem} \label{thm:dying-prob}
	For $\vec{\bm{N}}_L = (d_{\text{in}}, N,\cdots,N,d_{\text{out}}) \in \mathbb{N}^{L+1}$,
	let $\bm{\theta}_{\vec{\bm{N}}_L} = \{\bm{W}^\ell, \bm{b}^\ell\}_{1 \le \ell \le L}$
	and $\mathcal{N}^L(\x; \bm{\theta}_{\vec{\bm{N}}_L})$ be a ReLU neural network with $L$ layers, each having $N$ neurons.
	Suppose that all weights and biases
	are randomly initialized
	from probability distributions, which satisfy
	\begin{equation} \label{thm1:condition}
	    P\left( \bm{W}^\ell_j \in \mathbb{R}^{N_{\ell-1}}_{-}, \bm{b}^\ell_j < 0 \right) \ge p > 0, \qquad
	    \forall 1 \le j \le N_{\ell}, 
	    \forall 1 \le \ell \le L,
	\end{equation}
	for some constant $p > 0$, where $\bm{W}^\ell_j$ is the $j$-th row of the $\ell$-th layer weight matrix and $\bm{b}^\ell_j$ is the $j$-th component of the $\ell$-th layer bias vector.
	Then 
	\begin{equation*}
	\begin{split}
	\lim_{L\to \infty} P(\mathfrak{J}_{\vec{\bm{N}}_L})  = 1,
	\end{split}
	\end{equation*}
	where $\mathfrak{J}_{\vec{\bm{N}}_L}$
	is defined in \eqref{def:dyingNN-event}.
\end{theorem} 
\begin{proof}
    The proof can be found in Appendix~\ref{app:thm:dying-prob}.
\end{proof}

We remark that Equation~\ref{thm1:condition} is a very mild condition and it can be satisfied in many cases.
For example, when symmetric probability distributions around 0 are employed, the condition is met with $p = 2^{-N-1}$.
Theorem~\ref{thm:dying-prob} implies that the fully connected ReLU network will be dead at the initialization as long as the network is deep enough.
This explains theoretically why training a very deep network is hard.

Theorem~\ref{thm:dying-prob} shows that the ReLU network asymptotically will be dead. 
Thus, we are now concerned with the convergence behavior of the probability of NNs being BD.
Since almost all common initialization procedures use symmetric probability distributions around 0, 
we derive an upper bound of the born dead probability (BDP)
for symmetric initialization.

\begin{theorem} \label{thm:main}
	For $\vec{\bm{N}}_L = (d_{\text{in}},N_1, \cdots,N_{L-1},d_{\text{out}}) \in \mathbb{N}^{L+1}$,
	let $\bm{\theta}_{\vec{\bm{N}}_L} = \{\bm{W}^\ell, \bm{b}^\ell\}_{1 \le \ell \le L}$
	and
	$\mathcal{N}^L(\x; \bm{\theta}_{\vec{\bm{N}}_L})$ be a ReLU neural network with $L$ layers, each having $N_1, \cdots, N_{L-1}$ neurons.
	Suppose that all weights are independently initialized from symmetric probability distributions around 0 
	and all biases are either drawn from a symmetric distribution or set to zero.
	Then 
	\begin{equation} \label{eqn:sym-dying-prob-upp}
	P(\mathfrak{J}_{\vec{\bm{N}}_L})
	\le
	1 - \prod_{\ell=1}^{L-1}(1-(1/2)^{N_\ell}),
	\end{equation}
	where $\mathfrak{J}_{\vec{\bm{N}}_L}$
	is defined in \eqref{def:dyingNN-event}.
	Furthermore, 
	if $\vec{\bm{N}}_L = (d_{\text{in}},N,\cdots,N,d_{\text{out}}) \in \mathbb{N}^{L+1}$,
	\begin{equation*}
	\begin{split}
	\lim_{L\to \infty} P(\mathfrak{J}_{\vec{\bm{N}}_L})  = 1,
	\qquad
	\lim_{N\to \infty} P(\mathfrak{J}_{\vec{\bm{N}}_L})  = 0.
	\end{split}
	\end{equation*}
\end{theorem} 
\begin{proof}
    The proof can be found in Appendix~\ref{app:thm:main}.
\end{proof}

Theorem~\ref{thm:main} provides an upper bound of the BDP. It shows that at a fixed depth $L$, the network will not be BD in probability as the number of width $N$ goes to infinite.
In order to understand how this probability behaves with respect to the number of width and depth, 
a lower bound is needed.
We thus provide a lower bound
of the BDP of ReLU NNs at $d_{\text{in}} = 1$.
\begin{theorem} \label{thm:nnwidthN}
	For a network architecture $\vec{\bm{N}}_L = (1,N, \cdots,N,d_{\text{out}}) \in \mathbb{N}^{L+1}$,
	let $\bm{\theta}_{\vec{\bm{N}}_L} = \{\bm{W}^\ell, \bm{b}^\ell\}_{1 \le \ell \le L}$
	and $\mathcal{N}^L(\x; \bm{\theta}_{\vec{\bm{N}}_L})$ be a ReLU neural network with $L$ layers, each having $N$ neurons.
	Suppose that all weights are independently initialized from continuous symmetric probability distributions around 0, which satisfies 
	\begin{align*}
		P(\langle \bm{W}_j^\ell, \bm{v}_{1}\rangle > 0, \langle \bm{W}_j^\ell, \bm{v}_{2}\rangle < 0 | \bm{v}_{1},\bm{v}_2) \le \frac{1}{4}, \quad \bm{0} \ne \bm{v}_1, \bm{v}_2 \in \mathbb{R}_{+}^{N_{\ell-1}}, \quad \forall 1 \le j \le N_{\ell},
	\end{align*} 
	where $\bm{W}_j^\ell$ is the $j$-th row of the $\ell$-th layer weight matrix,
	and all biases are initialized to zero.
	Then
	\begin{equation*}
	p_{\text{low}}(L,N)
	\le
	P(\mathfrak{J}_{\vec{\bm{N}}_L})
	\le
	1 - \prod_{\ell=1}^{L-1}(1-(1/2)^{N}),
	\end{equation*}
	where
	$\mathfrak{J}_{\vec{\bm{N}}_L}$
	is defined in \eqref{def:dyingNN-event},
	$a_1 = 1 - (1/2)^N$, $a_2 = 1-(1/2)^{N-1}-(N-1)(1/4)^N$,
	and
    \begin{equation*}
    p_{\text{low}}(L,N) =
	1 - a_1^{L-2} + \frac{(1-2^{-N+1})(1-2^{-N})}{1+(N-1)2^{-N}}
	(-a_1^{L-2} + a_2^{L-2}).
    \end{equation*}
\end{theorem}
\begin{proof}
    The proof can be found in Appendix~\ref{app:thm:nnwidthN}.
\end{proof}

\begin{figure}[htbp]
\centering
\includegraphics{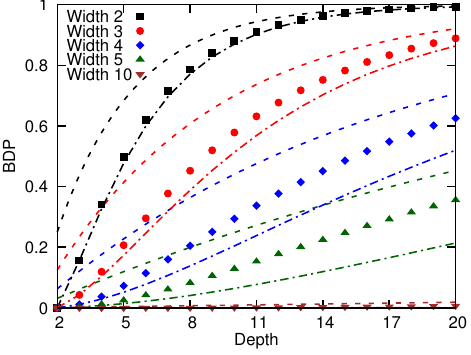}
\caption{Probability of a ReLU NN to be born dead as a function of the number of layers for different widths. The dash lines represent the upper and lower bounds from Theorem~\ref{thm:nnwidthN}. The symbols represent our numerical estimations. Similar colors correspond to the same width.
}\label{fig:p_zeroinit_din1}
\end{figure}

Theorem~\ref{thm:nnwidthN} reveals that the BDP behavior depends on the network architecture.
In Fig.~\ref{fig:p_zeroinit_din1}, we plot
the BDP with respect to increasing the number of layers at varying width from $N=2$ to $N=5$.
A bias-free ReLU feed-forward NN with $d_{\text{in}}=1$ is employed with weights randomly initialized from symmetric distributions. 
The results of one million independent simulations 
are used to calculate each probability estimation. Numerical estimations are shown as symbols.
The upper and lower bounds from Theorem~\ref{thm:nnwidthN} are also plotted with dash and dash-dot lines, respectively.
We see that when the NN gets narrower, the probability of NN being BD grows faster as 
the depth increases. 
Also, at a fixed width $N$, the BDP grows as the number of layer increases.
This is expected by Theorems~\ref{thm:dying-prob} and
\ref{thm:nnwidthN}.

Once the network is BD, 
we have no hope to train the network successfully.
Here we provide a formal statement of the consequence of the network being BD.
\begin{theorem}\label{thm:nn2mean}
For any $\vec{\bm{N}}_L$, 
suppose $\bm{\theta}_{\vec{\bm{N}}_L}$ is initialized to be in
$\mathfrak{J}_{\vec{\bm{N}}_L}$
defined in \eqref{def:dyingNN-event}.
Then, for any loss function $\mathcal{L}$, 
and for any gradient based method,
the ReLU network 
is optimized to be a constant function, 
which minimizes the loss.
\end{theorem}
\begin{proof}
    The proof can be found in Appendix~\ref{app:thm:nn2mean}.
\end{proof}

Theorem~\ref{thm:nn2mean} implies that no matter what gradient-based optimizers are employed including stochastic gradient desecent (SGD), SGD-Nesterov~\cite{sutskever2013importance}, AdaGrad~\cite{duchi2011adaptive}, AdaDelta~\cite{zeiler2012adadelta}, RMSProp~\cite{hintonlecture6a}, Adam~\cite{kingma2014adam}, BFGS~\cite{nocedal2006nonlinear}, L-BFGS~\cite{byrd1995limited}, 
the network is trained to be a constant function which minimizes the loss.

If the online-learning or the stochastic gradient method is employed, where the training data are independently drawn from a probability distribution $P_{\D}$, 
the optimized network is 
$$
\N^L(\x;\bm{\theta}^*) = \bm{c}^* = \argmin_{\bm{c} \in \R^{N_L}} 
\mathbb{E}
\left[ \ell(\bm{c}, f(\x))) \right],
$$
where the expectation $\mathbb{E}$ is taken with respect to $\x \sim P_{\D}$. For example, if $L^2$-loss is employed, i.e., $\ell(\N^L(\x),f(\x)) = (\N^L(\x)-f(\x))^2$, the resulting network is $\mathbb{E}[f(\x)]$.
If $L^1$ loss is employed, i.e., $\ell(\N^L(\x),f(\x)) = |\N^L(\x)-f(\x))|$, the resulting network is the median of $f(\x)$ with respect to $\x \sim P_\D$.
Note that the mean absolute error (MAE) and the mean squared error (MSE) used in practice are discrete versions of $L^1$ and $L^2$ loss, respectively, if the size of minibatch is large.

When we design a neural network, we want the BDP to be small, say, less than 1\% or 10\%. 
Then, the upper bound (Equation~\ref{eqn:sym-dying-prob-upp}) of Theorem~\ref{thm:main} can be used for designing a specific network architecture, which has a small probability of NNs being born dead.

\begin{corollary}\label{cor:p}
For fixed depth $L$,
let $\vec{\bm{N}}_L = (d_{\text{in}},N, \cdots,N, d_{\text{out}}) \in \mathbb{N}^{L+1}$.
For $\delta > 0$, suppose the width $N$ is $N = \log_2 L/\delta$.
Then,
$P(\mathfrak{J}_{\vec{\bm{N}}_L}^c) > 1 - \delta$,
where 
$\mathfrak{J}_{\vec{\bm{N}}_L}^c$ is the complement of $\mathfrak{J}_{\vec{\bm{N}}_L}$
defined in \eqref{def:dyingNN-event}.
\end{corollary}
\begin{proof}
    This readily follows from
\begin{align*}
    P(\mathfrak{J}_{\vec{\bm{N}}_L}) \le 1 - (1-2^{-N})^{L-1} 
    \le  1 - (1-(L-1)2^{-N}) \le L2^{-N} = \delta.
\end{align*}
\end{proof}

\begin{figure}[htbp]
\centering
\includegraphics{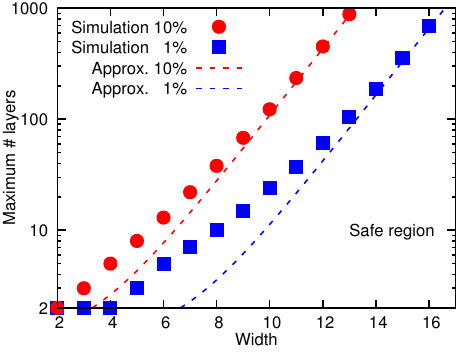}
\caption{Diagram indicating safe operating regions for a ReLU NN. The dash lines represent Corollary~\ref{cor:p} while the symbols represent our numerical tests. The maximum number of layers of a neural network can be used at different width to keep the probability of collapse less than 1\% or 10\%. The region below the blue line is the safe region when we design a neural network. As the width increases the theoretical predictions match closer with our numerical simulations.}\label{fig:max_layer}
\end{figure}

As a practical guide, we constructed a diagram shown in Fig.~\ref{fig:max_layer} that includes both theoretical predictions and our numerical tests. We see that as the number of layers increases, the numerical tests match closer to the theoretical results. It is clear from the diagram that a 10-layer NN of width 10 has a probability of dying less than 1\% whereas a 10-layer NN of width 5 has a probability of dying greater than 10\%; for width of three the probability is about 60\%.
Note that the growth rate of the maximum number of layers is exponential which is expected by Corollary~\ref{cor:p}.

Our results explain why deep and narrow networks have not been popularly employed
in practice and reveal a limitation of training extremely deep networks.
We remark that modern practical networks happen to be
sufficiently wide enough to have small BDPs.
We believe that 
the born dead network problem 
is a major impediment for one to explore 
a potential use of deep and narrow networks.

\section{Randomized asymmetric initialization} \label{sec:new-init}
The so-called `He initialization' \cite{he2015delving} is perhaps one of the most popular initialization schemes in deep learning community, especially when the ReLU activation function is concerned.
The effectiveness of the He initialization has been shown in many machine learning applications.
The He initialization uses mean zero normal distributions.
Thus, as we discussed earlier, it suffers from the dying ReLU.
We thus propose a new initialization procedure, namely, a randomized asymmetric initialization.
The motivation is in twofolds. One is to mimics the He initialization so that the new scheme can produce similar generalization performance. 
The other is to alleviate the problem of dying ReLU neural networks. We hope that the proposed initialization opens a new way to explore the potential use of deep and narrow networks in practice.
 
For ease of discussion, we introduce some notation.
For any vector $\bm{v} \in \mathbb{R}^{n+1}$ and $k \in \{1,\cdots, n+1\}$, we define
\begin{equation} \label{def-v_-k}
\bm{v}_{-k} = \left(v_1,\cdots, v_{k-1}, v_{k+1}, \cdots, v_{n+1}\right)^T \in \mathbb{R}^n.
\end{equation}
In order to train a $L$-layer neural network, we need to initialize $\bm{\theta}_L = \{\bm{W}^\ell, \bm{b}^\ell\}_{1\le \ell \le L}$.
At each layer, let $\textbf{V}^{\ell} = [\W^{\ell}, \bm{b}^{\ell}] \in \mathbb{R}^{N_\ell \times (N_{\ell-1}+1)}$.
We denote the $j$-th row of $\textbf{V}^{\ell}$ by $\textbf{V}^{\ell}_j = [\W^{\ell}_{j}, \bm{b}^{\ell}_j] \in \mathbb{R}^{N_{\ell-1}+1}$, $j=1,\cdots,N_\ell$
where $N_0 = d_{\text{in}}$ and $N_L = d_{\text{out}}$.

\subsection{Proposed initialization} \label{subsec:asyminit}
We propose to initialize $\textbf{V}^{\ell}$ as follows.
Let $\text{P}_\ell$ be a probability distribution defined on $[0,M_\ell]$ for some $M_\ell > 0$ or $[0,\infty)$.
Note that $\text{P}_\ell$ is asymmetric around 0.
At the first layer of $\ell=1$, we employ the `He initialization' \cite{he2015delving}, i.e., 
$\textbf{W}^1_{ij} \sim N(0,2/d_{\text{in}})$
and $\textbf{b}^1 = \textbf{0}$.
For $\ell \ge 2$, and each $1\le j \le N_{\ell}$, we initialize $\bm{V}_j^\ell$ as follows:
\begin{enumerate}
	\item Randomly choose $k^\ell_j$ in $\{1,2,\cdots, N_{\ell-1}, N_{\ell-1}+1\}$.
	\item Initialize $(\textbf{V}^{\ell}_{j})_{-k^\ell_j} \sim \mathcal{N}(0,\sigma_\ell^2\bm{I})$
	and $(\textbf{V}^{\ell}_{j})_{k^\ell_j} \sim \text{P}_\ell$.
\end{enumerate} 
Since an index is randomly chosen at each $\ell$ and $j$
and a positive number is randomly drawn from an asymmetric probability distribution around 0,
we name this new initialization a randomized asymmetric initialization.
Only for the first layer, the He initialization is employed.
This is because since an input could have a negative value, if the weight which corresponds to the negative input were to be initialized from $\text{P}_\ell$, this could cause the dying ReLU.
We note that the new initialization requires us to choose $\sigma_\ell^2$ and $\text{P}_\ell$. 
In Subsection~\ref{subsec:length-map}, these will be theoretically determined.
One could choose multiple indices in the step 1 of the new initialization. However, for simplicity, we constraint ourselves to a single index case. 

We first show that this new initialization procedure results in a smaller upper bound of the BDP.
\begin{theorem} \label{thm:asym-prob}
	For $\vec{\bm{N}}_L = (d_{\text{in}},N_1, \cdots,N_{L-1},d_{\text{out}}) \in \mathbb{N}^{L+1}$,
	let $\bm{\theta}_{\vec{\bm{N}}_{L}} = \{\bm{W}^\ell, \bm{b}^\ell\}_{1 \le \ell \le L}$
	and
	$\mathcal{N}^L(\x; \bm{\theta}_{\vec{\bm{N}}_L})$ be a ReLU neural network with $L$ layers, each having $N_1, \cdots, N_{L-1}$ neurons.
    Suppose $\bm{\theta}_{\vec{\bm{N}}_L}$
	is initialized by the randomized asymmetric initialization (RAI).
	Then,
	\begin{equation*}
	P(\mathfrak{J}_{\vec{\bm{N}}_L}) 
	\le 
	1 - \prod_{\ell=1}^{L-1} \left(1 -
	\left(1/2 - \gamma_{\ell}\right)^{N_{\ell}}\right),
	\end{equation*}
	where $\mathfrak{J}_{\vec{\bm{N}}_L}$ is defined in \eqref{def:dyingNN-event},
	$\gamma_1 = 0$ and $\gamma_j$'s are some constants in $(0,0.5]$, which depend on $\{N_\ell\}_{\ell=1}^{L-1}$ and the training input data $\D$.
\end{theorem}
\begin{proof}
    The proof can be found in Appendix~\ref{app:thm:asym-prob}.
\end{proof}

When a symmetric initialization is employed, 
$\gamma_j = 0$ for all $1\le j < N_L$,
which results in Equation~\ref{eqn:sym-dying-prob-upp} of Theorem~\ref{thm:main}.
Although the new initialization has a smaller upper bound compared to those by symmetric initialization,
as Theorem~\ref{thm:dying-prob} suggests, 
it also asymptotically suffers from the dying ReLU.

\begin{corollary} \label{cor:RAI-p}
	Assuming the same conditions in Theorem~\ref{thm:asym-prob}, and 
	$N_\ell = N$ for all $\ell$.
	Let $\mathfrak{J}_{\vec{\bm{N}}_L}$
    be the event defined in \eqref{def:dyingNN-event}.
	Then, there exists $2 < \gamma$, which depends on $N, L$ and the training input data $\D$, such that 
	\begin{equation*}
	P(\mathfrak{J}_{\vec{\bm{N}}_L}) 
	\le 
	1 - \prod_{\ell=1}^{L-1} \left(1 -
	(1/\gamma)^{N}\right).
	\end{equation*}
	For fixed depth $L$ and $\delta > 0$, if the width $N$ is $N = \log_\gamma L/\delta$, 
	we have
    $P(\mathfrak{J}_{\vec{\bm{N}}_L}^c) > 1 - \delta$,
    where 
    $\mathfrak{J}_{\vec{\bm{N}}_L}^c$ is the complement of $\mathfrak{J}_{\vec{\bm{N}}_L}$.
	Furthermore, 
	\begin{equation*}
	\begin{split}
	\lim_{L\to \infty} 
	P(\mathfrak{J}_{\vec{\bm{N}}_L}) 
	= 1,
	\qquad
	\lim_{N\to \infty} 
	P(\mathfrak{J}_{\vec{\bm{N}}_L}) 
	= 0.
	\end{split}
	\end{equation*}
\end{corollary}
\begin{proof}
    The proof is readily followed from Theorem~\ref{thm:dying-prob},~\ref{thm:asym-prob} and
    Corollary~\ref{cor:p}.
\end{proof}

\subsection{Second moment analysis} \label{subsec:length-map}
The proposed randomized asymmetric initialization (RAI) described in Subsection~\ref{subsec:asyminit} requires us to determine 
$\sigma_\ell^2$ and $\text{P}_\ell$.
Similar to the He initialization (\cite{he2016deep}), 
we aim to properly choose initialization parameters from the length map analysis.
Following the work of~\cite{poole2016exponential}, we present the analysis of a single input propagation through the deep ReLU network.
To be more precise, 
for a network architecture $\vec{\bm{N}}_{\ell} = (d_{\text{in}},N_1, \cdots,N_{\ell-1},N_{\ell}) \in \mathbb{N}^{\ell+1}$,
we track the expectation of the normalized 
squared length of the input vector at each layer,
$\mathbb{E}[q^\ell(\x;\bm{\theta}_{\vec{\bm{N}}_{\ell}})]$,
where 
$q^\ell(\x;\bm{\theta}_{\vec{\bm{N}}_{\ell}}) = \frac{\|\N^\ell(\x;\bm{\theta}_{\vec{\bm{N}}_{\ell}})\|^2}{N_\ell}$.
The expectation $\mathbb{E}$ is taken with respect to all weights and biases $\bm{\theta}_{\vec{\bm{N}}_{\ell}}$.

\begin{theorem} \label{thm:2ndmo-asyminit}
    For $\vec{\bm{N}}_{\ell} = (d_{\text{in}},N_1, \cdots,N_{\ell-1},N_{\ell}) \in \mathbb{N}^{\ell+1}$,
    let $\bm{\theta}_{\vec{\bm{N}}_{\ell}} = \{\bm{W}^j, \bm{b}^j\}_{1 \le j \le \ell}$ be initialized by the RAI
    with the following  
    $\sigma_{\ell}^2$ and $\text{P}_\ell$.
    Let $\sigma_\ell^2 := \frac{\sigma_w^2}{N_{\ell-1}}$
    for some $\sigma_w^2$
    and let $\text{P}_\ell$ be a probability distribution 
    whose support is $[0, M_\ell] \subset \mathbb{R}^+$
    such that 
    $\mu_{\ell,i}'=E[X_\ell^i] < \infty$, for $i=1,2$,
    where $X_\ell \sim \text{P}_\ell$ 
    and $\mu_{\ell,1}' \ge M_\ell/2$.
	Then for any input $\x \in \mathbb{R}^{d_{in}}$, we have
	\begin{align*}
	\frac{\mathcal{A}_{low, \ell}}{2}\mathbb{E}[q^{\ell}(\x;\bm{\theta}_{\vec{\bm{N}}_{\ell}})]
	+ \sigma_{b,\ell}^2
	\le 
	\mathbb{E}[q^{\ell+1}(\x;\bm{\theta}_{\vec{\bm{N}}_{\ell+1}})]
	\le 
	\frac{\mathcal{A}_{upp, \ell}}{2} \mathbb{E}[q^{\ell}(\x;\bm{\theta}_{\vec{\bm{N}}_{\ell}})]
	+ \sigma_{b,\ell}^2,
	\end{align*}
	where
	$\sigma_{b,\ell}^2 = \frac{\mu_{\ell+1,2}'+
	\sigma_{\ell+1}^2N_{\ell+1}}{N_{\ell+1}+1}$, $\mathcal{A}_{low, \ell} = 
	\frac{\sigma_{b,\ell+1}^2}{\sigma_{b,\ell}^2}
	\left(\frac{N_\ell\mu_{\ell,2}'+N_{\ell-1}\sigma_w^2}{N_{\ell-1}+1}\right)$, and 
	$$\mathcal{A}_{upp, \ell} = 
	\frac{\sigma_{b,\ell+1}^2}{\sigma_{b,\ell}^2}\left(
	\frac{N_{\ell-1}\sigma_w^2 +2\sqrt{2/\pi}N_{\ell}\mu_{\ell,1}'\sigma_w+ 2N_{\ell}\mu_{\ell,2}'}{N_{\ell-1}+1}\right).$$
\end{theorem}
\begin{proof}
    The proof can be found in Appendix~\ref{app:thm:2ndmo-asyminit}.
\end{proof}

\begin{corollary} \label{cor:qell}
	Under the same conditions of 
	Theorem~\ref{thm:2ndmo-asyminit},
	if $N_\ell = N$, $M_\ell = M$, $E[X_{\ell}^i] = \mu_i'$, $i=1,2$ for all $\ell$,
	and $\mu_1' \ge M/2$, 
	we have
	\begin{align*}
	\frac{\mathcal{A}_{low, \ell}}{2}\mathbb{E}[q^{\ell}(\x;\bm{\theta}_{\vec{\bm{N}}_{\ell}})]
	+ \sigma_{b,\ell}^2
	\le 
	\mathbb{E}[q^{\ell+1}(\x;\bm{\theta}_{\vec{\bm{N}}_{\ell+1}})]
	\le 
	\frac{\mathcal{A}_{upp, \ell}}{2} \mathbb{E}[q^{\ell}(\x;\bm{\theta}_{\vec{\bm{N}}_{\ell}})]
	+ \sigma_{b,\ell}^2,
	\end{align*}
	where
	$
	\sigma_{b,\ell}^2 = \frac{\mu_{2}'
	+\sigma_{w}^2}{N+1}$, $\mathcal{A}_{low, \ell} = 
	\frac{N(\mu_{2}'+\sigma_w^2)}{N+1}$, 
	and 
	$\mathcal{A}_{upp, \ell} = 
	\frac{N(\sigma_w^2 + 2\sqrt{2/\pi}\mu_1'\sigma_w + 2\mu_2')}{N+1}$.
\end{corollary}

Since $\sigma_{b,\ell}^2 > 0$, $\lim_{\ell \to \infty} \mathbb{E}[q^{\ell}(\x;\bm{\theta}_{\vec{\bm{N}}_{\ell}})]$ cannot be zero.
In order for $\lim_{\ell \to \infty} \mathbb{E}[q^{\ell}(\x;\bm{\theta}_{\vec{\bm{N}}_{\ell}})] < \infty$, the initialization parameters $(\sigma_w^2, \mu_{\ell,1}', \mu_{\ell,2}')$ have to be chosen to satisfy
$\mathcal{A}_{upp, \ell} < 2$.
Assuming $\mu_2' < 1$, if $\sigma_w$ is chosen to be 
\begin{equation} \label{def:sigmW}
    \sigma_w = \sqrt{2}\left(-\frac{\mu_1'}{\sqrt{\pi}} + \sqrt{\frac{\mu_1'^2}{\pi}+1-\mu_2'}\right),
\end{equation}
we have $\frac{\mathcal{A}_{upp, \ell}}{2}= \frac{N}{N+1}$
which satisfies the condition.

\begin{figure}[htbp]
\centering
\includegraphics[width=7.4cm]{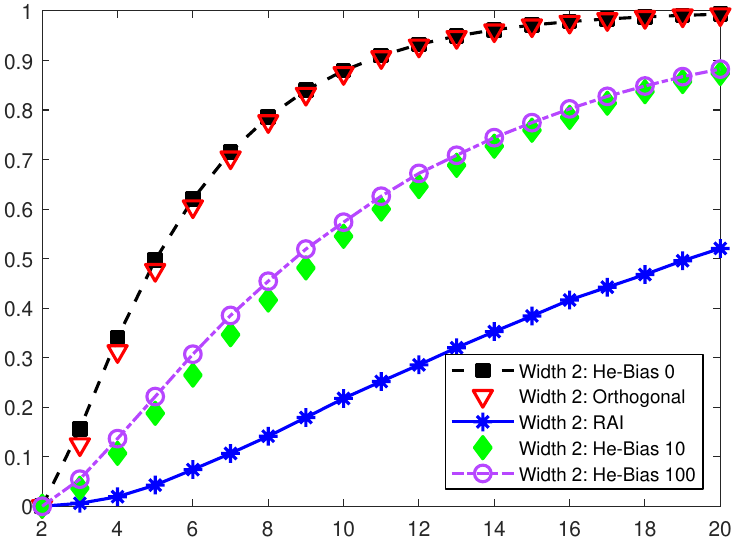}
\includegraphics[width=7.4cm]{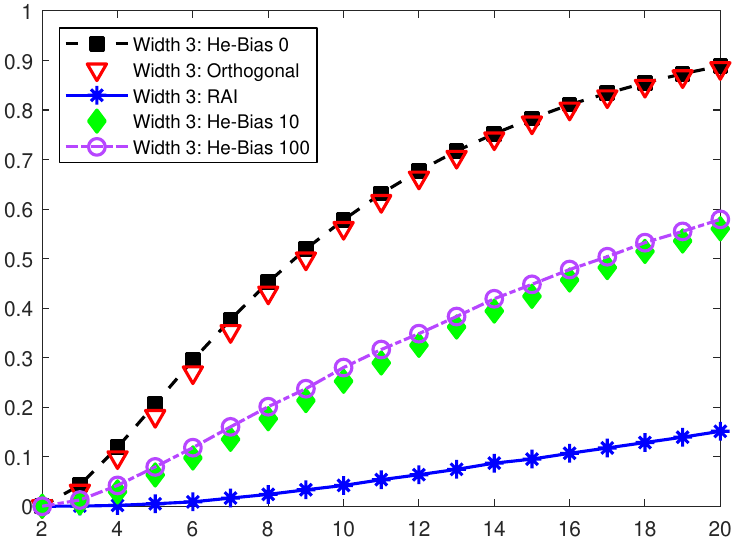}
\includegraphics[width=7.4cm]{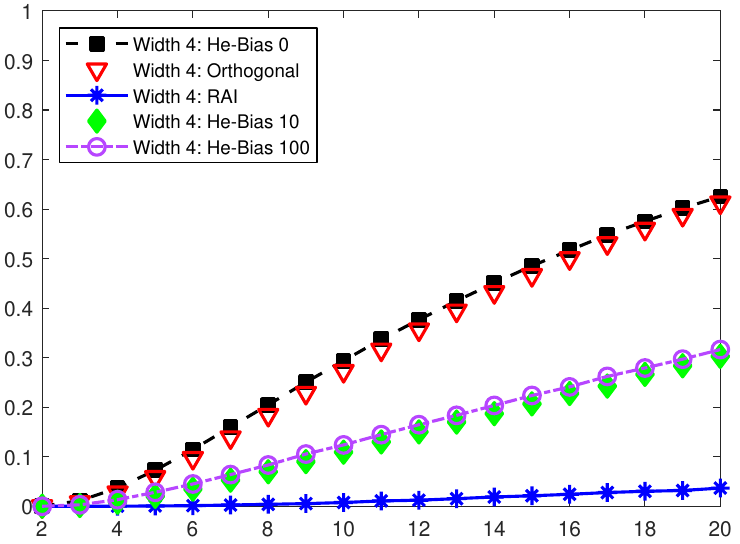}
\includegraphics[width=7.4cm]{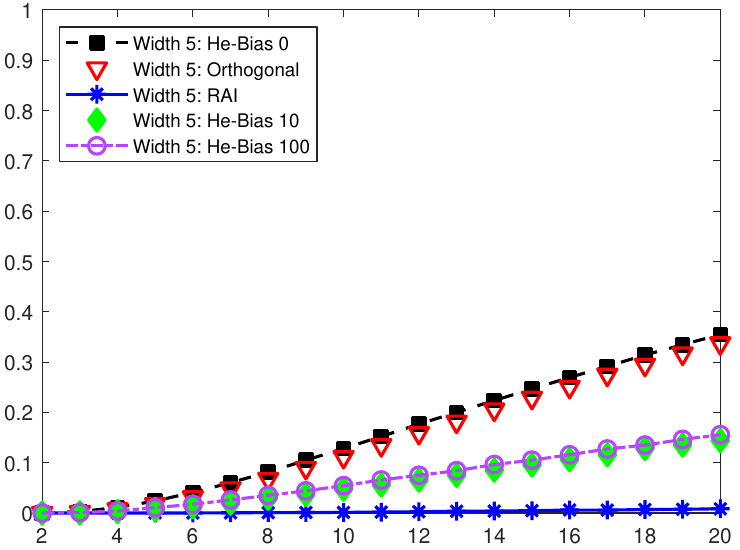}
\caption{The BDPs are plotted with respect to increasing the number of depth $L$ at varying width $N=2$ (top left), $N=3$ (top right), $N=4$ (bottom left) and $N=5$ (bottom right).
The ReLU neural networks in $d_{\text{in}}=1$
are employed.
The square, diamond and circle symbols correspond to the He initialization \cite{he2015delving} with constant bias 0, 10 and 100, respectively. 
The inverted triangle symbols correspond to the orthogonal initialization \cite{saxe2013exact}.
The asterisk symbols correspond to the proposed randomized asymmetric initialization (RAI).
}\label{fig:orthogonal}
\end{figure}

\begin{figure}[htbp]
\centering
\includegraphics{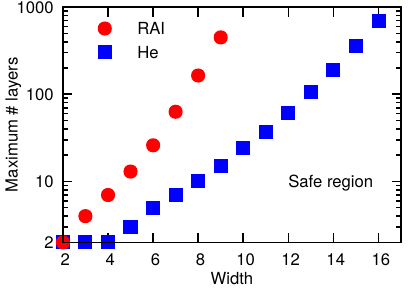}
\caption{Comparison of the safe operating regions for a ReLU NN between RAI and He initializations. The maximum number of layers of a neural network can be used at different width to keep the probability of collapse less than 1\%.}\label{fig:max_layer_RAI}
\end{figure}

\subsection{Comparison against other initialization procedures}
In Fig.~\ref{fig:orthogonal}, we demonstrate the probability that the network is BD by the proposed randomized asymmetric initialization (RAI) method.
Here we employ $\text{P}=\text{Beta}(2,1)$ and $\sigma_w=-\frac{2\sqrt{2}}{3\sqrt{\pi}} + \sqrt{1+\frac{8}{9\pi}} \approx 0.6007$ from Equation~\ref{def:sigmW}
(see Appendix~\ref{app:RAI_code} for the Python code).
To compare against other procedures, we present 
the results by the He initialization \cite{he2015delving}.
We also present the results of existing asymmetric initialization procedures; the orthogonal~\cite{saxe2013exact} and the layer-sequential unit-variance (LSUV)~\cite{mishkin2015all} initializations.
The LSUV is the orthogonal initialization combined with rescaling of weights such that the output of each layer has unit variance. Because weight rescaling cannot make the output escape from the negative part of ReLU, it is sufficient to consider the orthogonal initialization.
We see that the BDPs by the orthogonal initialization are very close to and a little lower than those by the He initialization. 
This implies that the orthogonal initialization cannot prevent the dying ReLU network.
Furthermore, we show the results by the He initialization with positive constant bias of 10 and 100. 
Naively speaking, having a big positive bias will help in preventing dying ReLU neurons, as the input of each layer is pushed to be positive, although this might cause the exploding gradient problem in training.
We see that the BDPs by the He with bias 10 and 100
are lower than those by the He with bias 0 and the orthogonal initialization.
However, it is clearly observed that 
our proposed initialization (RAI) drastically drops the BDPs compared to all others.
This is implied by Theorem~\ref{thm:asym-prob}.

As a practical guide, we constructed a diagram shown in Fig.~\ref{fig:max_layer_RAI} to demonstrate the 1\% safe operation region by the RAI.
It is clear from the diagram that 
the RAI drastically expands its 1\% safe region
compared to those by the He initialization.
Note that the growth rate of the maximum number of layers is exponential which is expected by Corollary~\ref{cor:RAI-p}.
\section{Numerical examples} \label{sec:example}
We demonstrate the effectiveness of the proposed randomized asymmetric initialization (RAI) in training deep ReLU networks. 

Test functions include one- and two-dimensional functions of different regularities. The following test functions are employed as 
unknown target functions. For one dimensional cases, 
\begin{equation} \label{test-func}
    \begin{split}
        f_1(x) = |x|, \qquad
        f_2(x) = x\sin(5x), \qquad
        f_3(x) = 1_{\{x>0\}}(x) + 0.2\sin(5x).
    \end{split}
\end{equation}
For two dimensional case,
\begin{equation} \label{test-func-2d}
    f_4(x_1,x_2) = \begin{bmatrix} |{x}_1 + {x}_2 | \\ |{x}_1 - {x}_2 | \end{bmatrix}. 
\end{equation}
We employ the network architecture having the width of $d_{\text{in}} + d_{\text{out}}$ at all layers.
Here $d_{\text{in}}$ and $d_{\text{out}}$ are the dimensions of the input and output, respectively.
It was shown in \cite{hanin2017approximating}
that the minimum number of width required for the universal approximation
is less than or equal to $d_{\text{in}} + d_{\text{out}}$.
We thus choose this specific network architecture.
as it theoretically guarantees to approximate any continuous function.
In all numerical examples, 
we employ one of the most popular first-order gradient-based optimization, \texttt{Adam} \cite{kingma2014adam} with the default parameters.
The minibatch size is chosen to be either 64 or 128.
The standard $L_2$-loss function $\ell(\N^L(\x,\bm{\theta}),(\x,y)) = (\N^L(\x;\bm{\theta}) - y)^2$ is used on 3,000 training data.
The training data are randomly uniformly drawn from $[-\sqrt{3},\sqrt{3}]^{d_{\text{in}}}$.
Without changing any setups described above, we present the approximation results based on different initialization procedures.
The results by our proposed randomized asymmetric initialization are referred to `Rand. Asymmetric' or `RAI'. 
Specifically, we use $\text{P} = \text{Beta}(2,1)$ with $\sigma_w$ defined in Equation~\ref{def:sigmW}.
To compare against other methods, we also show the results by the He initialization \cite{he2015delving}.
We present the ensemble of 1,000 independent training simulations.

In one dimensional examples, we employ a 10-layer ReLU network of width 2. 
It follows from Fig.~\ref{fig:orthogonal}
that 
we expect to observe at least 88\% training results by the symmetric initialization
and 22\% training results by the RAI
are collapsed.
In the two dimensional example, 
we employ a 20-layer ReLU network of width 4.
According to Fig.~\ref{fig:orthogonal},
we expect to see at least 63\% training results by the symmetric initialization
and 3.7\% training results by the RAI
are collapsed.

\begin{figure}[htbp]
\centering
\includegraphics{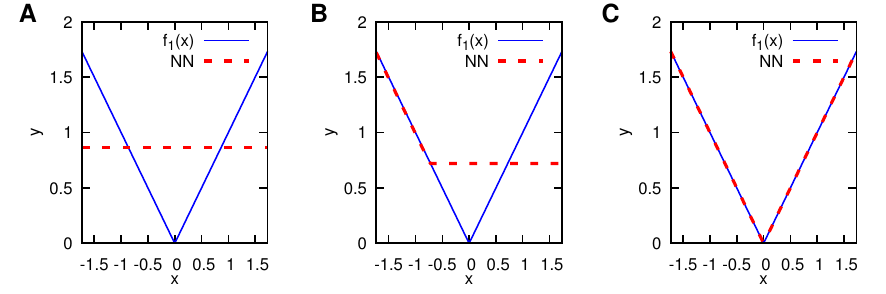}
\scalebox{1}{
\begin{tabular}{|>{\centering}m{4.0cm}|>{\centering}m{2.55cm}|>{\centering}m{2.6cm}|>{\centering}m{2.55cm}|@{}m{0pt}@{}}
 \hline
 $ \small \bm{f_1(x)}$ 
 & \small \textbf{Collapse (A)} 
 & \small \textbf{Half-Trained (B)} 
 & \small \textbf{Success (C)} 
 &\\[10pt]
 \hline 
 \small Symmetric (He init.) 
 & 93.6\%
 & 4.2\% 
 & 2.2\%
 &\\
 \small Rand. Asymmetric   
 & \textbf{40.3\%} 
 & \textbf{22.4\%}  
 & \textbf{37.3\%} 
 &\\
 \hline
\end{tabular}}
\caption{The approximation results for $f_1(x)$ using a 10-layer ReLU network of width 2. 
For this specific test function, we observe only 3 trained results shown in \text{A, B, C}. 
The table shows the corresponding empirical probabilities from 1,000 independent simulations.
The only difference is the initialization.
}
\label{fig:abs}
\end{figure}

Fig.~\ref{fig:abs} shows all training outcomes of our 1,000 simulations for approximating $f_1(x)$ and its corresponding empirical probabilities by different initialization schemes.
For this specific test function, we observe only 3 trained results shown in \text{A, B, C}.
In fact, $f_1(x)$ can be represented exactly by a 2-layer ReLU network with width 2, $f_1(x) = |x| = \text{ReLU}(x)+\text{ReLU}(-x)$.
It can clearly be seen that the He initialization results in the collapse with probability more than 90\%.
However, this probability is drastically reduced to 40\% by the RAI.
These probabilities are different from the probability that the network is BD. 
This implies that even though the network wasn't BD, there are cases that after training, the network dies.
In this example, 5.6\% and 18.3\% of results by the symmetric and our method, respectively, are not dead at the initialization, however, they are ended up with collapsing after training.
The 37.3\% of training results by the RAI perfectly recover the target function $f_1(x)$, however,
only 2.2\% of results by the He initialization achieve this success.
Also, 22.4\% of the RAI and 4.2\% of the He initialization produce the half-trained results which correspond to Fig.~\ref{fig:abs} (B). We remark that the only difference in training is the initialization. 
This implies that our new initialization does not only prevent the dying ReLU network but also improves the quality of the training in this case.

\begin{figure}[htbp]
\centering
\includegraphics{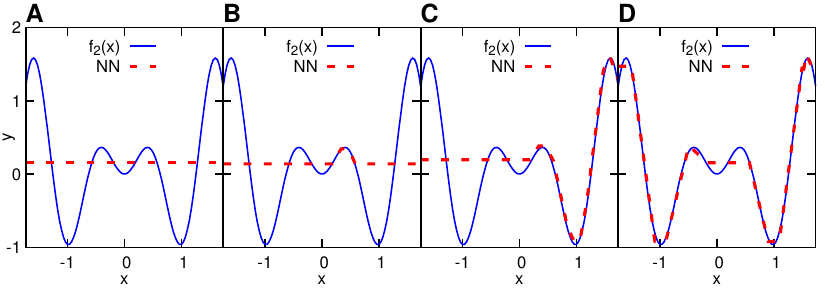}
\scalebox{1}{
\begin{tabular}{|>{\centering}m{4.0cm}|>{\centering}m{4.05cm}|>{\centering}m{4.05cm}|@{}m{0pt}@{}} 
 \hline
 $ \small \bm{f_2(x)}$ 
 & \small \textbf{Collapsed (A)} 
 & \small \textbf{Not collapsed (B,C,D)} 
 &\\
 \hline 
 \small Symmetric (He init.) 
 & 91.9\%
 & 8.1\%  
 &\\
 \small Rand. Asymmetric   
 & \textbf{29.2\%} 
 & \textbf{70.8\%}  
 &\\
 \hline
\end{tabular}}
\caption{
The approximation results for $f_2(x)$ using a 10-layer ReLU network of width 2. 
Among many trained results, four are shown. 
The table shows the corresponding empirical probabilities from 1,000 independent simulations.
The only difference is the initialization.
}
\label{fig:xsin5x}
\end{figure}

The approximation results for $f_2(x)$ are shown in Fig.~\ref{fig:xsin5x}. 
Note that $f_2$ is a $C^{\infty}$ function.
It can be seen that
91.9\% of training results by the symmetric initialization
and 29.2\% of training results by the RAI
are collapsed which correspond to Fig.~\ref{fig:xsin5x} (A).
This indicates that the RAI can effectively alleviate the dying ReLU.
In this example, 3.9\% and 7.2\% of results by the symmetric and our method, respectively, are not dead at the initialization, however, they are ended up with collapsing after training.
Except for the collapse, other training outcomes are not easy to be classified.
Fig.~\ref{fig:xsin5x} (B,C,D) shows three training results among many others.
We observe that the behavior and result of training are not easily predictable in general. 
However, we consistently observe partially collapsed results after training.
Such partial collapses are also observed in Fig.~\ref{fig:xsin5x} (B,C,D).
We believe that this requires more attention and 
postpone the study of this partial collapse to future work.

\begin{figure}[htbp]
\centering
\includegraphics{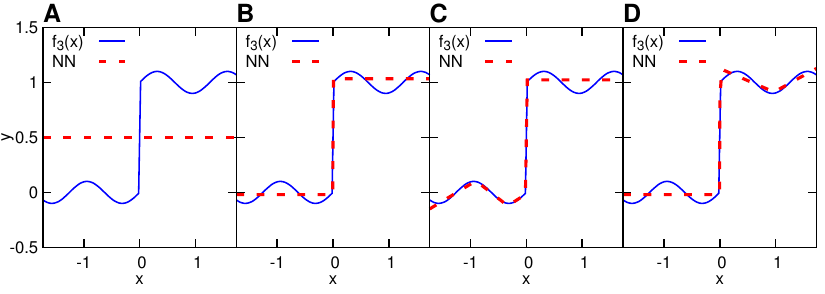}
\scalebox{1}{
\begin{tabular}{|>{\centering}m{4.0cm}|>{\centering}m{4.05cm}|>{\centering}m{4.05cm}|@{}m{0pt}@{}} 
 \hline
 $\small \bm{f_3(x)}$ 
 & \small \textbf{Collapsed (A)} 
 & \small \textbf{Not collapsed (B,C,D)} 
 &\\
 \hline 
 \small Symmetric (He init.) 
 & 93.8\%
 & 6.2\%  
 &\\
 \small Rand. Asymmetric   
 & \textbf{32.6\%} 
 & \textbf{67.4\%}  
 &\\
 \hline
\end{tabular}}
\caption{
The approximation results for $f_3(x)$ using a 10-layer ReLU network of width 2. 
Among many trained results, four are shown. 
The table shows the corresponding empirical probabilities from 1,000 independent simulations.
The only difference is the initialization.
}
\label{fig:stepsin}
\end{figure}

Similar behavior is observed for approximating a discontinuous function $f_3(x)$.
The approximation results for $f_3(x)$ and its corresponding empirical probabilities are shown in 
Fig.~\ref{fig:stepsin}.
We see that
93.8\% of training results by the He initialization
and 32.6\% of training results by the RAI
are collapsed which correspond to Fig.~\ref{fig:stepsin} (A).
In this example, the RAI drops the probability of collapsing by 60.3 percentage point.
Again, this implies that the RAI can effectively avoid the dying ReLU, especially when deep and narrow ReLU networks are employed.
Fig.~\ref{fig:stepsin} (B,C,D) shows three trained results among many others.
Again, we observe partially collapsed results.

\begin{figure}[htbp]
\centering
\includegraphics{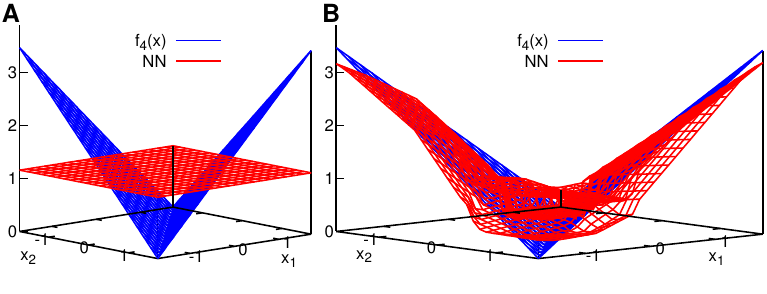}
\scalebox{1}{
\begin{tabular}{|>{\centering}m{4.0cm}|>{\centering}m{4.05cm}|>{\centering}m{4.05cm}|@{}m{0pt}@{}} 
 \hline
 $\small \bm{f_4(x)}$ 
 &\small \textbf{Collapsed (A)} 
 &\small \textbf{Not collapsed (B)} 
 &\\
 \hline 
 \small Symmetric (He init.) 
 & 76.8\%
 & 23.2\%  
 &\\
 \small Rand. Asymmetric   
 & \textbf{9.6\%} 
 & \textbf{90.4\%}  
 &\\
 \hline
\end{tabular}}
\caption{
The approximation results for $f_4(\x)$ using a 20-layer ReLU network of width 4. 
Among many trained results, two are shown. 
The table shows the corresponding empirical probabilities from 1,000 independent simulations.
The only difference is the initialization.
}
\label{fig:abs2d}
\end{figure}

Next we show the approximation result for a multi-dimensional inputs and outputs function $f_4(\x)$ defined in Equation~\ref{test-func-2d}.
We observe similar behavior.
Fig.~\ref{fig:abs2d} shows some of approximation results for $f_4$ and its corresponding probabilities.
Among 1,000 independent simulations, the collapsed results are obtained by the He initialization with 76.8\% probability
and by the RAI with 9.6\% probability. 
From Fig.~\ref{fig:orthogonal}, we expect to observe at least 63\% and 3.7\% of results by the symmetric and the RAI to be collapsed.
Thus, in this example, 13.8\% and 5.9\% of results by the symmetric and our method, respectively, are not dead at the initialization, however, they are ended up with Fig.~\ref{fig:abs2d} (A) after training.
This indicates that the RAI can also effectively overcome the dying ReLU in multi-dimensional inputs and outputs tasks.

\begin{figure}[htbp]
    \centering
    \includegraphics{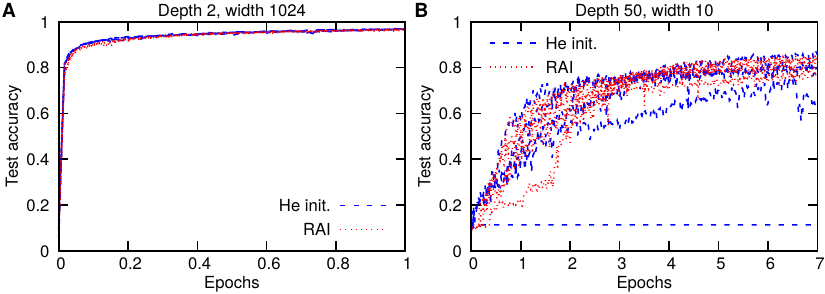}
    \caption{
    The test accuracy on the MNIST of five independent simulations are shown with respect to the number of epochs
    by the He initialization and the RAI.
    (Left) A shallow (depth 2, width 1024) ReLU network is employed. The He initialization and the RAI have similar performance. 
    (Right) A deep (depth 50, width 10) ReLU network is employed.
    The RAI results in higher test accuracy than the He initialization.}
    \label{fig:mnist}
\end{figure}

As a last example, we demonstrate the performance of the RAI on the MNIST dataset. 
For the training, we employ the cross-entropy loss 
and the mini-batch size of 100.
The networks are trained using \texttt{Adam} \cite{kingma2014adam} with its default values.
In Fig. \ref{fig:mnist}, the convergence of the test accuracy is shown with respect to the number of epochs.
On the left and right,
we employ the ReLU network of depth 2 and width 1024
and of depth 50 and with 10, respectively.
We see that when the shallow and wide network is employed, the RAI and the He initialization show similar generalization performance (test accuracy).
However, when the deep and narrow network is employed,
the RAI performs better than the He initialization.
This indicates that the proposed RAI not only reduces the BDP of deep networks, but also has good generalization performance.

\section{Conclusion} \label{sec:conclusion}

In this paper, 
we establish, to the best of our knowledge, the first theoretical analysis on the dying ReLU.
By focusing on the worst case of dying ReLU,
we define `the dying ReLU network' which refers to 
the problem when the ReLU network is dead.
We categorize the dying process into two phases.
One phase is the event where the ReLU network is initialized to be a constant function.
We refer to this event as `the network is born dead'.
The other phase is the event where the ReLU network is collapsed after training.
Certainly, the first phase implies the second, but not vice versa.
We show that the probability that the network is born dead
goes to 1 as the depth goes infinite.
Also, we provide an upper and a lower bound of the dying probability in $d_{\text{in}}=1$ when the standard symmetric initialization is used.

Furthermore, in order to overcome the dying ReLU networks, we propose a new initialization procedure, namely, a randomized asymmetric initialization (RAI).
We show that the RAI has a smaller upper bound of the probability of NNs being born dead.
By establishing the expected length map relation (second moment analysis),
all parameters needed for the new method are theoretically designed.
Numerical examples are provided to demonstrate the performance of our method.
We observe that the RAI does not only overcome the dying ReLU but also improves the training and generalization performance.

\section*{Acknowledgements}

This work received support by the DARPA EQUiPS grant N66001-15-2-4055, the AFOSR grant FA9550-17-1-0013, and the DARPA AIRA grant HR00111990025. The research of the third author was partially supported by the NSF of China 11771083 and the NSF of Fujian 2017J01556, 2016J01013.

\appendix

\section{Proof of Theorem~\ref{thm:dying-prob}} \label{app:thm:dying-prob}
The proof starts with the following lemma.
\begin{lemma} \label{app:lemma:nndying}
Let $\N^L(\x)$ be a $L$-layer ReLU neural network with $N_\ell$ neurons at the $\ell$-th layer.
Suppose all weights are randomly independently generated from probability distributions satisfying $P(\bm{W}_j^\ell \bm{z} = \bm{0}) = 0$ for any nonzero vector $\bm{z} \in \mathbb{R}^{N_{\ell-1}}$ and any $j$-th row of $\bm{W}^\ell$.
Then 
\begin{equation*}
    P(\mathfrak{J}_{\vec{\bm{N}}_L})
    = P(\exists \hspace{0.1cm} \ell \in \{1,\dots,L-1\} \text{ such that } \phi(\N^\ell(\x)) = \bm{0} \hspace{0.1cm} \forall \x \in \D),
\end{equation*}
where $\D \subset B_r(\bm{0})=\{\x \in \mathbb{R}^{d_{\text{in}}} | \|x\| < r\}$ for any $r>0$
and $\mathfrak{J}_{\vec{\bm{N}}_L}$
is the event defined in \eqref{def:dyingNN-event}.
\end{lemma}
\begin{proof}
    Suppose $\N^L(\x) = \N^L(\bm{0})$ for all $\x \in \D \subset B_r(\bm{0})$. Then $\phi(\N^{L-1}(\x)) = \phi(\N^{L-1}(\bm{0}))$ for all $\x \in \D$.
    If $\phi(\N^{L-1}(\x)) = \phi(\N^{L-1}(\bm{0})) = \bm{0}$,
    we are done as $\ell = L-1$.
    If it is not the case, 
    there exists $j$ in $\{1,\cdots, N_{L-1}\}$ such that 
    for all $\x \in \D$,
    $$
    (\N^{L-1}(\x))_j = \bm{W}^{L-1}_j\phi(\N^{L-2}(\x)) + \bm{b}^{L-1}_j = \bm{W}^{L-1}_j\phi(\N^{L-2}(\bm{0})) + \bm{b}^{L-1}_j = (\N^{L-1}(\bm{0}))_j > 0.
    $$
    Thus we have $\bm{W}^{L-1}_j
    \left(\phi(\N^{L-2}(\x))-\phi(\N^{L-2}(\bm{0}))\right) = 0$
    for all $\x \in \D$.
    Let consider the following events:
    \begin{align*}
        &G_{L-1}:=\{\bm{W}^{L-1}_j
    \phi(\N^{L-2}(\x))=
    \bm{W}^{L-1}_j\phi(\N^{L-2}(\bm{0})), \forall \x \in \D \}, \\
    &R_{L-2} := \{
    \phi(\N^{L-2}(\bm{x})) = \phi(\N^{L-2}(\bm{0})), \forall \x \in \D \}.
    \end{align*}
    Note that 
    $P(G_{L-1}|R_{L-2}) = 1$. Also, since 
    $P(\bm{W}^\ell \bm{z}) = 0$ for any nonzero vector $\bm{z}$, we have
    $P(G_{L-1}|R_{L-2}^c) = 0$.
    Therefore, 
    \begin{align*}
        P(G_{L-1}) = P(G_{L-1}|R_{L-2})P(R_{L-2}) +  P(G_{L-1}|R_{L-2}^c)P(R_{L-2}^c) = P(R_{L-2}).
    \end{align*}
    Thus we can focus on
    $\phi(\N^{L-2}(\x)) = \phi(\N^{L-2}(\bm{0})), \forall \x \in \D$.
    If $\phi(\N^{L-2}(\x)) = \phi(\N^{L-2}(\bm{0})) = \bm{0}$,
    we are done as $\ell = L-2$.
    If it is not the case, it follows from the similar procedure 
    that  
    $\phi(\N^{L-3}(\x)) = \phi(\N^{L-3}(\bm{0}))$ in $\D$.
    By repeating these, we conclude that 
    $$
    P(\mathfrak{J}_{\vec{\bm{N}}_L})
    = P(\exists \hspace{0.1cm} \ell \in \{1,\dots,L-1\} \text{ such that } \phi(\N^\ell(\x))) = \bm{0} \hspace{0.1cm} \forall \x \in \D).
    $$
\end{proof}

\begin{proof}
Let $\D \subset B_r(0) \subset \mathbb{R}^{d_{\text{in}}}$ be a training domain where $r$ is any positive real number.
	We consider a probability space $(\Omega, \mathcal{F}, P)$
	where all random weight matrices and bias vectors are defined on.
	For every $\ell \ge 1$, let $\mathcal{F}_\ell$ be a sub-$\sigma$-algebra of $\mathcal{F}$
	generated by $\{\bm{W}^j, \bm{b}^j\}_{1\le j \le \ell}$.
	Since $\mathcal{F}_k \subset \mathcal{F}_\ell$ for $k \le \ell$, 
	$(\mathcal{F}_\ell)$ is a filtration.
	Let us define the events of our interest $\{A_\ell\}_{2 \le \ell}$ where
	\begin{equation} \label{app:def:A_ell}
	\begin{split}
	    A_\ell &:= 
	    \mathfrak{J}_{\vec{\bm{N}}_{\ell}}
	\overset{a.s.}{=} \{\exists \hspace{0.1cm} j \in \{1,\dots,\ell-1\} \text{ such that } \phi(\N^j(\x)) = \bm{0} \hspace{0.1cm} \forall \x \in \D\},
	\end{split}
	\end{equation}
	where the second equality is from Lemma~\ref{app:lemma:nndying}.
	Note that $A_\ell$ is measurable in $\mathcal{F}_{\ell-1}$.
	Here $\{\bm{b}^{\ell}\}_{1 \le \ell}$ could be either $0$ or random vectors.
	Since $\N^1(\x) = \bm{W}^1\x + \bm{b}^1$, $P(A_1) = 0$.
	To calculate $P(A_\ell)$ for $\ell \ge 2$, let consider another event 
	$C_{\ell,k}$ on which exactly $(N_\ell-k)$-components of $\phi(\N^\ell(\x))$ are zero on $\D$.
	For notational completeness, we set $C_{1,k} = \emptyset$ for $0 \le k < N_1$
	and $C_{1,N_1} = \Omega$.
	Then since $C_{\ell-1, 0} \subset A_\ell$, we have
    \begin{equation} \label{app:thm1:eqn2}
     P(A_\ell) \ge P(C_{\ell-1,0}).   
    \end{equation}
	We want to show $\lim_{\ell \to \infty} P(C_{\ell,0}) = 1$.
	Since $\{C_{\ell-1,k}\}_{0\le k \le N_{\ell-1}}$ is a partition of $\Omega$,
	by the total law of probability, we have
	\begin{align*}
	P(C_{\ell,s}) &= \sum_{k=0}^{N_{\ell-1}} P(C_{\ell,s}|C_{\ell-1,k}) P(C_{\ell-1,k}),
	\end{align*}
	where 
	$P(C_{\ell,0}|C_{\ell-1,0}) = 1$, 
	and 
	$P(C_{\ell,s}|C_{\ell-1,0}) = 0, \forall s \ge 1$.
	Since $\W_{ij}^\ell$ and $\bm{b}^\ell_j$ are independently initialized, we have
	\begin{align*}
	    P(C_{\ell,0}|C_{\ell-1,k}) = 
	    \left(
	    P( \langle \bm{W}^{\ell}_j, \phi(\N^{\ell-1}(\x)) \rangle + \bm{b}^{\ell}_j < 0 |C_{\ell-1,k}) 
	    \right)^{N_{\ell}} \ge p_k^{N_{\ell}} > 0,
	\end{align*}
	where the second and the third inequalities hold from the assumption. 
	Here $p_k > 0$ does not depend on $\ell$.
	If $\bm{W}_{ij}^\ell, \bm{b}_j^\ell$ are randomly initialized from symmetric distributions around 0, 
	\begin{align*}
	P(C_{\ell,0}|C_{\ell-1,k}) \ge 
	\begin{cases}
	(2^{-k})^{N_{\ell}} & \text{if $\bm{b}^{\ell} = 0$} \\
	(2^{-(k+1)})^{N_{\ell}}  & \text{if $\bm{b}^{\ell}$ is generated from a symmetric distribution}
	\end{cases}
	\end{align*}

	Let define a transition matrix $V_\ell$ of size $(N_{\ell-1}+1) \times (N_{\ell}+1)$ such that the $(i+1,j+1)$-component is defined to be
	\begin{align*}
	V_\ell(i+1,j+1) = P(C_{\ell,j}|C_{\ell-1,i}), \qquad \text{where} \quad
	0 \le j \le N_{\ell} \quad \text{and} \quad 0 \le i \le N_{\ell-1}.  
	\end{align*}
	Then given 
	$$\pi_1 = [P(C_{1,0}),P(C_{1,1}),\cdots, P(C_{1,N_1})] = [0,\cdots, 0,1] \in \mathbb{R}^{N_1+1},$$ 
	we have
	\begin{equation*}
	\pi_{\ell} = \pi_1 V_2 \cdots V_\ell = [P(C_{\ell,0}),P(C_{\ell,1}),\cdots, P(C_{\ell,N_{\ell}})], \qquad \ell \ge 2.
	\end{equation*}
	
	Suppose $N_\ell = N$ for all $\ell \ge 1$.
	Note that the first row of $V_\ell$ is $[1,0,\cdots, 0]$ for all $\ell \ge 2$.
	Thus we have the following strictly increasing sequence $\{a_\ell\}_{\ell=1}^\infty$:
	$$
	a_\ell := (\pi_\ell)_1 = P(C_{\ell,0}).
	$$
	Since $a_\ell \le 1$, it converges, say, $\lim_{\ell \to \infty} a_\ell = a \le 1$.
	Suppose $a \ne 1$, i.e., $a-1 < 0$
	and let $p_* = \min_{1\le k \le N} p_k$.
	Then since $a_{k+1} = (\pi_k V_{k+1})_1$,
	we have
	\begin{align*}
	a_{k+1} = a_k + \sum_{j =1}^{N} P(C_{k+1,0}|C_{k,j}) (\pi_k)_{j+1} \ge a_k + (1-a_k)(p_*^{-(N+1)})^N.
	\end{align*}
	Thus
	\begin{align*}
	0 \le a - a_{k+1} \le a - a_k + (a_k - 1)(p_*^{-(N+1)})^N.
	\end{align*}
	By taking limit on the both sides, we have
	\begin{align*}
	0 \le (a-1)(p_*^{-(N+1)})^N < 0,
	\end{align*}
	which leads a contradiction.
	Therefore, $a = \lim_{\ell \to \infty} P(C_{\ell,0}) = 1$.
	It then follows from Equation~\ref{app:thm1:eqn2} that
	$$
	\lim_{\ell \to \infty} 
	P(\mathfrak{J}_{\vec{\bm{N}}_{\ell}})
	\ge
	\lim_{\ell \to \infty} P(C_{\ell-1,0}) = 1,
	$$
	which completes the proof.
\end{proof}

\section{Proof of Theorem~\ref{thm:main}}
\label{app:thm:main}

\begin{proof}
Based on Lemma~\ref{app:lemma:nndying}, let consider
	\begin{align*}
	    A_\ell &= \{\exists \hspace{0.1cm} j \in \{1,\dots,\ell-1\} \text{ such that } \phi(\N^\ell(\x))) = \bm{0} \hspace{0.1cm} \forall \x \in \D\}, \\ 
	    A_\ell^c &= 
	    \{\text{$\forall 1 \le j < \ell$ there exists $\x \in \D$ such that } \phi(\N^{j}(\x)) \ne \bm{0}\},\\
	    \tilde{A}_{\ell,\x}^c &= \{\forall 1 \le j < \ell, \hspace{0.1cm} \phi(\N^{j}(\x)) \ne \bm{0} \}, \\
	    \tilde{A}_{\ell,\x} &= \{\exists \hspace{0.1cm} j \in \{1,\dots,\ell-1\} \text{ such that } \phi(\N^{j}(\x)) = \bm{0} \}.
	\end{align*}
	Then if $\x \in \D$, $\tilde{A}_{\ell,\x}^c \subset A_\ell^c$.
	Thus it suffices to compute $P(\tilde{A}_{\ell,\x}^c)$
	as 
	$$
	P(A_\ell) = 1-P(A_\ell^c) \le 1-P(\tilde{A}_{\ell,\x}^c).
	$$
    For $\x \ne \bm{0}$, let consider
    \begin{align*}
        U_{j,\x} = \{\phi(\N^j(\x)) = \bm{0} \}, \qquad
        U_{j,\x}^c = \{\phi(\N^j(\x)) \ne \bm{0} \}.
    \end{align*}
    Note that $\bigcup_{1\le j < \ell} U_{j,\x} = \tilde{A}_{\ell,\x}$
    and  
    $\tilde{A}_{\ell,\x}^c = \bigcap_{1\le j < \ell} U_{j,\x}^c$.
    Since 
    $P(\tilde{A}_{j,\x}^c|\tilde{A}_{j-1,\x}) = 0$ for all $j$,
    we have
    \begin{equation} \label{app:thm1:eqn3}
        P(\tilde{A}_{\ell,\x}^c)
        = P(\tilde{A}_{\ell,\x}^c | \tilde{A}_{\ell-1,\x}^c)P(\tilde{A}_{\ell-1,\x}^c)
        = \cdots 
        = P(\tilde{A}_{1,\x}^c)\prod_{j=2}^\ell P(\tilde{A}_{j,\x}^c | \tilde{A}_{j-1,\x}^c).
    \end{equation}
    Note that $P(\tilde{A}_{1,\x}^c)=1$.
    Also, note that since the rows of $\bm{W}^\ell$ and $\bm{b}_j^\ell$ are independent,
    \begin{equation} \label{app:thm1:eqn4-prob}
        P(\tilde{A}_{j,\x} |\tilde{A}_{j-1,\x}^c) 
        = \prod_{s=1}^{N_{j-1}} 
        P\left(\bm{W}^{j-1}_s\phi(\N^{j-2}(\x)) + \bm{b}^{j-1}_s \le 0 |\tilde{A}_{j-1,\x}^c\right). 
    \end{equation}
    Since the weight and biases are randomly drawn from symmetry distribution around 0 and \\
    $P\left(\bm{W}^{j-1}_s\phi(\N^{j-2}(\x)) + \bm{b}^{j-1}_s = 0 |\tilde{A}_{j-1,\x}^c\right) = 0$,
    we obtain
    $$
    P\left(\bm{W}^j_s\phi(\N^{j-1}(\x)) + \bm{b}^j_s \le 0 |\tilde{A}_{j-1,\x}^c\right) = \frac{1}{2}.
    $$
    Therefore, 
    $P(\tilde{A}_{j,\x}|\tilde{A}_{j-1,\x}^c)=2^{-N_{j-1}}$
    and thus,
    $P(\tilde{A}_{j,\x}^c|\tilde{A}_{j-1,\x}^c)=1-2^{-N_{j-1}}$.
    It then follows from Equation~\ref{app:thm1:eqn3} that
    \begin{equation*}
        P(\tilde{A}_{\ell,\x}^c) = \prod_{j=1}^{\ell-1} (1-2^{-N_{j}}),
    \end{equation*}
    which completes the proof.
\end{proof}

\section{Proof of Theorem~\ref{thm:nnwidthN}} \label{app:thm:nnwidthN}
\begin{proof}
    We now assume $d_{\text{in}}=1$, $N_\ell = N$ and 
	without loss of generality let $\D \subset [-r,r]$ be a training domain for any $r>0$.
	Also we assume that all weights are initialized from continuous symmetric probability distributions around 0.
	And the biases are set to zeros.
	
	
	Since $d_{\text{in}}= 1$, for each $\ell$, there exist 
	non-negative vectors $\bm{v}_{\pm} \in \mathbb{R}^{N_\ell}_{+}$ such that
	\begin{equation} \label{app:thm2:l-th-layer}
	\phi(\N^\ell(x)) = \bm{v}_{+} \phi(x) + 
	\bm{v_{-}}\phi(-x).
	\end{equation}
	Let $B_{\ell,0}$ be the event where
	$\phi(\N^\ell(x)) = 0$,
	and let 
	$B_{\ell,1}$ be the event where
	$$
	\phi(\N^\ell(x)) = \bm{v}_{+} \phi(x), \quad \text{or} \quad 
	\bm{v_{-}}\phi(-x),
	\quad \text{or} \quad 
	\bm{v}(\phi(x) + b\phi(-x))
	$$
	for some $\bm{v}_\pm$, $\bm{v} \in \mathbb{R}_+^{N_\ell}$ and $b > 0$.
	Let $B_{\ell,2}$ be the event where
	$$
	\phi(\N^\ell(x)) = \bm{v}_{+} \phi(x) +
	\bm{v_{-}}\phi(-x)
	$$
	for some linearly independent vectors $\bm{v}_{\pm}$.
	Then it can be checked that $P(B_{\ell+1,k}|B_{\ell,s}) = 0$ for all $2 \ge k > s \ge 0$.
	Thus, it suffices to consider 
	$P(B_{\ell+1,k}|B_{\ell,s})$ where 
	$0 \le k \le s \le 2$.
	At $\ell=1$, since $\phi(\N^1(\x)) = \phi(\W^{1}\x)$ and $\D \subset [-r,r]$,
	we have 
	\begin{align*}
	P(B_{1,0}) &= 0, \qquad
	P(B_{1,1}) = 2^{-N+1}, \qquad	P(B_{1,2}) = 1-2^{-N+1}.
	\end{align*}
	For $\ell > 1$, 
	it can be checked that 
	$P(B_{\ell,0}|B_{\ell-1,0}) = 1$, 
	$P(B_{\ell,0}|B_{\ell-1,1}) = 2^{-N}$,
	and thus $P(B_{\ell,1}|B_{\ell-1,1}) = 1-2^{-N}$.
	For $P(B_{\ell,0}|B_{\ell-1,2})$ and $P(B_{\ell,1}|B_{\ell-1,2})$, 
	we observe the followings.	
	In $B_{\ell,2}$, we have
	\begin{align*}
	\phi(\langle \bm{w},\phi(\N^\ell(x)) \rangle) = 
	\phi(\langle \bm{w}, \bm{v}_{+}\rangle) \phi(x)
	+
	\phi(\langle \bm{w}, \bm{v}_{-}\rangle) \phi(-x),.
	\end{align*}
	Since $\bm{v}_{\pm}$ are
	nonzero vectors in $\mathbb{R}^{N_\ell}_+$
	and thus satisfy $\langle \bm{v}_+, \bm{v}_- \rangle \ge 0$, 
	by the assumption, we obtain
	\begin{align*}
		P(\langle \bm{w}, \bm{v}_{+}\rangle < 0, \langle \bm{w}, \bm{v}_{-}\rangle < 0|\bm{v}_{\pm}) &= \frac{1}{2} - p_{\bm{v}_\pm} \ge 1/4, \\
		P(\langle \bm{w}, \bm{v}_{+}\rangle > 0, \langle \bm{w}, \bm{v}_{-}\rangle > 0 | \bm{v}_{\pm}) &= \frac{1}{2} - p_{\bm{v}_\pm} \ge 1/4, \\
		P(\langle \bm{w}, \bm{v}_{+}\rangle > 0, \langle \bm{w}, \bm{v}_{-}\rangle < 0 | \bm{v}_{\pm}) &=
		P(\langle \bm{w}, \bm{v}_{+}\rangle < 0, \langle \bm{w}, \bm{v}_{-}\rangle > 0 | \bm{v}_{\pm}) =p_{\bm{v}_\pm} \le \frac{1}{4}.
	\end{align*} 
	Thus we have
	\begin{align*}
	P(B_{\ell,0}|B_{\ell-1,2}) = \mathbb{E}\left[ \left(\frac{1}{2} - p_{\bm{v}_\pm} \right)^N \right] \ge (1/4)^N
	\end{align*}
	where the expectation is taken under $p_{\bm{v}_\pm}$.
	For $P(B_{\ell,1}|B_{\ell-1,2})$,
	there are only three ways to move from $B_{\ell-1,2}$ to $B_{\ell,1}$. That is
	\begin{align*}
	   B_{\ell,1}^{(A)}|B_{\ell-1,2}: \phi(\N^{\ell-1}(x)) &\to  \phi(\N^{\ell}(x)) = \bm{v}_{+} \phi(x), \\
	   B_{\ell,1}^{(B)}|B_{\ell-1,2}: \phi(\N^{\ell-1}(x)) &\to
	     \phi(\N^{\ell}(x)) =\bm{v_{-}}\phi(-x), \\
	B_{\ell,1}^{(C)}|B_{\ell-1,2}: \phi(\N^{\ell-1}(x)) &\to
	 \phi(\N^{\ell}(x)) =\bm{v}(\phi(x) + b\phi(-x)).
	\end{align*}
	Thus 
	$
	P(B_{\ell,1}|B_{\ell-1,2}) = 
	P(B_{\ell,1}^{(A)}|B_{\ell-1,2})
	+
	P(B_{\ell,1}^{(B)}|B_{\ell-1,2})
	+
	P(B_{\ell,1}^{(C)}|B_{\ell-1,2}).
	$
	Note that due to the symmetry, 
	$P(B_{\ell,1}^{(A)}|B_{\ell-1,2}) = P(B_{\ell,1}^{(B)}|B_{\ell-1,2})$.
	Thus we have
	\begin{align*}
	P(B_{\ell,1}|B_{\ell-1,2}) &=
	2\left(\sum_{j =1}^N \binom{N}{j}\mathbb{E}\left[\left(\frac{1}{2} - p_{\bm{v}_\pm}\right)^{N-j}\left(p_{\bm{v}_\pm}\right)^j\right] \right) + \binom{N}{1}\mathbb{E}\left[\left(\frac{1}{2} - p_{\bm{v}_\pm}\right)^N\right] \\
	&= 2^{-N+1} + (N-2)P(B_{\ell,0}|B_{\ell-1,2}) \\
	&\ge 2^{-N+1} + (N-2)4^{-N}.
	\end{align*}
	Note that
	$$
	P(B_{\ell,0}|B_{\ell-1,2}) + P(B_{\ell,1}|B_{\ell-1,2})
	= 2^{-N+1} + (N-1)P(B_{\ell,0}|B_{\ell-1,2}).
	$$
	Since $P(A_{\ell+1}) = P(B_{\ell,0})$
	where $A_{\ell}$ is defined in Equation~\ref{app:def:A_ell}, 
	we aim to estimate $P(B_{\ell,0})$.
	Let $V_\ell$ be the transition matrix 
	whose $(i+1,j+1)$-component is $P(B_{\ell,j}|B_{\ell-1,i})$.
	Then $P(B_{\ell,0}) = \pi_1 V_2 \cdots V_\ell$
	where $(\pi_1)_j = P(B_{1,j-1})$.
 	By letting 
 	$$
 	\gamma_\ell 
 	= 1+\frac{2^{-N}- P(B_{\ell,0}|B_{\ell-1,2})}{2^{-N} + (N-1)P(B_{\ell,0}|B_{\ell-1,2})}.
 	$$
 	we obtain
 	\begin{align*}
 	V_\ell = Q_\ell D_\ell Q_\ell^{-1}, \qquad
 	Q_\ell = \begin{bmatrix}
 	1/\sqrt{3} & 0 & 0 \\
 	1/\sqrt{3} & \frac{1}{\sqrt{1+\gamma_\ell^2}} & 0 \\
 	1/\sqrt{3} & \frac{\gamma_\ell}{\sqrt{1+\gamma_\ell^2}}& 1
 	\end{bmatrix},
 	\quad
 	Q_\ell^{-1} =
 	\begin{bmatrix}
 	\sqrt{3} & 0 & 0 \\
 	-\sqrt{1+\gamma_\ell^2} & \sqrt{1+\gamma_\ell^2} & 0 \\
 	-(1-\gamma_\ell) & -\gamma_\ell & 1
 	\end{bmatrix}
 	\end{align*}
 	where $D_\ell = \text{diag}(V_\ell)$.
 	
	To find a lower bound of $P(B_{\ell,0})$, 
	we consider the following transition matrix $\mathcal{P}$ of size $3 \times 3$ which is defined to be 
	\begin{align*}
	\mathcal{P} = 
	\begin{bmatrix}
	1 & 0 & 0 \\
	(1/2)^N & 1 - (1/2)^N & 0 \\
	(1/4)^{N} & (1/2)^{N-1}+(N-2)(1/4)^N & 1-(1/2)^{N-1}-(N-1)(1/4)^N
	\end{bmatrix}.
	\end{align*}
	It can be checked that
	\begin{align*}
	\mathcal{P} = QDQ^{-1}, \qquad
	Q = \begin{bmatrix}
	1/\sqrt{3} & 0 & 0 \\
	1/\sqrt{3} & \frac{1}{\sqrt{1+\gamma^2}} & 0 \\
	1/\sqrt{3} & \frac{\gamma}{\sqrt{1+\gamma^2}}& 1
	\end{bmatrix},
	\quad
	Q^{-1} =
	\begin{bmatrix}
	\sqrt{3} & 0 & 0 \\
	-\sqrt{1+\gamma^2} & \sqrt{1+\gamma^2} & 0 \\
	-(1-\gamma) & -\gamma & 1
	\end{bmatrix}
	\end{align*}
	where 
	$\gamma = \frac{\mathcal{P}_{32}}{\mathcal{P}_{22} -\mathcal{P}_{33}}
	= \frac{2^{-N+1}+(N-2)4^{-N}}{2^{-N}+(N-1)4^{-N}} = 1 + \frac{2^{-N}-4^{-N}}{2^{-N}+(N-1)4^{-N}}$
	and $D = \text{diag}(\mathcal{P})$.
	Thus we have
	\begin{equation*}
	\mathcal{P}^\ell = 
	\begin{bmatrix}
	1 & 0 & 0 \\
	1 - (\mathcal{P}_{22})^\ell & (\mathcal{P}_{22})^\ell & 0 \\
	1-(\mathcal{P}_{22})^\ell-(\gamma-1)((\mathcal{P}_{22})^\ell-(\mathcal{P}_{33})^\ell) & \gamma((\mathcal{P}_{22})^\ell-(\mathcal{P}_{33})^\ell) & (\mathcal{P}_{33})^\ell
	\end{bmatrix}.
	\end{equation*}
	
	Similarly, we obtain 
 	\begin{equation*}
 	V_2\cdots V_{\ell+1} = 
 	\begin{bmatrix}
 	1 & 0 & 0 \\
 	1 - (\mathcal{P}_{22})^\ell & (\mathcal{P}_{22})^\ell & 0 \\
 	\xi_{\ell,31} & \xi_{\ell,32} & \xi_{\ell,33}
 	\end{bmatrix}
 	\end{equation*}
 	where $g(x) = 1 - 2^{-N+1} - (N-1)x$,
 	$p_\ell = (V_{\ell})_{31} = P(B_{\ell,0}|B_{\ell-1,2})$, 
 	$\gamma_{\ell,i} = \gamma_i$ for $1 \le i \le \ell$, 
 	$\gamma_{\ell,\ell+1} = 1$, 
 	\begin{align*}
 	\xi_{\ell, 31} &= (\mathcal{P}^\ell)_{31} -(\gamma_{\ell,1}-1)(g(p_1))^\ell 
 	+\sum_{i=1}^\ell (\gamma_{\ell,i} - \gamma_{\ell,i+1})(\mathcal{P}_{22})^{\ell-i}
 	\prod_{j=1}^i g(p_j),
 	\\
 	\xi_{\ell, 33} &= \prod_{j=1}^\ell g(p_j), \qquad
 	\xi_{\ell, 32} = 1 - \xi_{\ell, 31} - \xi_{\ell, 33}.
 	\end{align*}

	We want to show that 
	$$
	(\pi_1\mathcal{P}^{\ell})_1 
	\le (\pi_1 V_2 \cdots V_{\ell+1})_1 = P(B_{\ell+1,0})
	$$
	where
	\begin{align*}
	\pi_1 = 
	[P(B_{1,0}), P(B_{1,1}), P(B_{1,2})]
	= [0, 2^{-N+1}, 1 - 2^{-N+1} ].
	\end{align*}
	Let denote $\bar{\gamma_{\ell,i}}:= \gamma_{\ell,i}-1$
	and $g_i := g(p_i)$.
	It then suffices to show that 
	\begin{align*}
	\mathcal{J}:= \sum_{i=1}^\ell (\bar{\gamma_{\ell,i}} - \bar{\gamma_{\ell,i+1}})(\mathcal{P}_{22})^{\ell-i}
	\prod_{j=1}^i g_j
	-\bar{\gamma_{\ell,1}}(g_1)^\ell
	 \ge 0.
	\end{align*}
	Note that $4^{-N}=p_1 \le p_j <2^{-N}$, $\mathcal{P}_{22} > g(p_j)$,
	and thus
	$\mathcal{P}_{22}^\ell > (g(p_1))^\ell  \ge \prod_{j=1}^\ell g(p_j)$. 
	Also,
	\begin{align*}
	\mathcal{P}_{22} - g(p_i) = 2^{-N} + (N-1)p_i, \qquad
	\bar{\gamma_{\ell,i}} = \frac{2^{-N}-p_i}{2^{-N} + (N-1)p_i} = \frac{2^{-N}-p_i}{\mathcal{P}_{22} - g(p_i)}.
	\end{align*}
	Thus we have
	\begin{align*}
	\mathcal{J} &= \bar{\gamma_{\ell,1}}(g_1\mathcal{P}_{22}^{\ell-1} - g_1^\ell)
	- 
	\sum_{i =2}^\ell \bar{\gamma_{\ell,i}} (\mathcal{P}_{22} - g_i)
	\mathcal{P}_{22}^{\ell-i} \prod_{j=1}^{i-1} g_j  \\
	&\ge \bar{\gamma_{\ell,1}}(\mathcal{P}_{22}^\ell - g_1^\ell)
	- \sum_{i=1}^\ell \bar{\gamma_{\ell,i}} (\mathcal{P}_{22} - g_i)
	\mathcal{P}_{22}^{\ell-i} g_1^{i-1} \\
	&\ge \bar{\gamma_{\ell,1}}(\mathcal{P}_{22}^\ell - g_1^\ell)
	- \bar{\gamma_{\ell,1}}(\mathcal{P}_{22} - g_1)\frac{\mathcal{P}_{22}^\ell - g_1^\ell}{\mathcal{P}_{22} - g_1} = 0.
	\end{align*}
	Therefore, 
	$$
	(\mathcal{P}^\ell)_{31} \le (V_2\cdots V_{\ell+1})_{31},
	$$
	which implies that
	$$
	(\pi_1\mathcal{P}^{\ell})_1 
	\le (\pi_1 V_2 \cdots V_{\ell+1})_1 = P(B_{\ell+1,0}) = P(A_{\ell+2}).
	$$
	Furthermore, 
	it can be checked that 
	\begin{align*}
	(\pi \mathcal{P}^{\ell})_1 = 
	1 - (\mathcal{P}_{22})^\ell - \left(\frac{(1-2^{-N})(1-2^{-N+1})}{1+(N-1)2^{-N}}\right)
	((\mathcal{P}_{22})^\ell - (\mathcal{P}_{33})^\ell).
	\end{align*}
\end{proof}

\section{Proof of Theorem~\ref{thm:nn2mean}} \label{app:thm:nn2mean}
\begin{proof}
Since $\mathcal{N}^L(\mathbf{x})$ is a constant function,
it follows from Lemma~\ref{app:lemma:nndying} that
with probability 1,
there exists $\ell$ such that $\phi(\N^\ell(\x)) \equiv \mathbf{0}$.
Then the gradients of the loss function with respect to the weights 
and biases in the $1,\cdots, \ell$-th layers vanish.
Hence, the weights and biases in layers $1, \dots, \ell$ will not change when a gradient based optimizer is employed. This implies that $\mathcal{N}^L(\x)$ always remains to be a constant function as $\phi(\N^\ell(\x)) \equiv \mathbf{0}$.
Furthermore, the gradient method changes only the weights and biases in layers $\ell+1, \dots, L$.
Therefore, the ReLU NN can only be optimized to a constant function, which minimizes the loss function $\mathcal{L}$. 
\end{proof}

\section{Proof of Theorem~\ref{thm:asym-prob}} \label{app:thm:asym-prob}

\begin{lemma} \label{app:lemma:prob-positive-real}
	Let $\bm{v} \in \mathbb{R}^{n+1}$ be a vector such that
	 $\bm{v}_k \sim \text{P}$ and
	$\bm{v}_{-k} \sim N(0,\sigma^2\bm{I}_n)$
	where $\bm{v}_{-k}$ is defined in Equation~\ref{def-v_-k}.
	For any nonzero vector $\x \in \mathbb{R}^{n+1}$ whose $k$-th element is positive (i.e., $\x_k > 0$), let 
	\begin{equation*} 
	\|\tilde{\x}_{-k}\|^2 = \frac{\sum_{j \ne k} \x_j^2}{\x_k^2}, \qquad
	\tilde{\sigma}^2 = \|\tilde{\x}_{-k}\|^2\sigma^2.
	\end{equation*}
	Then 
	\begin{equation} \label{app:lemma:eqn:prob-positive}
	P\left(\langle \bm{v}, \x \rangle > 0 \right) =
	\begin{cases}
	\frac{1}{2} + \int_0^{M}
	(1-F_{\text{P}}(t))
	\frac{1}{\sqrt{2\pi}\tilde{\sigma}}e^{-\frac{t^2}{2\tilde{\sigma}^2}} dt & \text{if $\tilde{\sigma}^2 > 0$ and $\x_k > 0$}  \\
	1/2 & \text{if $\|{\x}_{-k}\|^2 > 0$ and $\x_k = 0$} \\
	1   & \text{if $\tilde{\sigma}^2 = 0$ and $\x_k > 0$},
	\end{cases}
	\end{equation}
	where $F_{\text{P}}(t)$ is the cdf of ${\text{P}}$.
\end{lemma}
\begin{proof}
	We first recall some properties of the normal distribution.
	Let $Y_1, \cdots, Y_n$ be i.i.d. random variables from $N(0,\sigma^2)$
	and let $\bm{Y}_n = (Y_1, \cdots, Y_n)$.
	Then for any vector $\bm{a} \in \mathbb{R}^n$,
	$$ 
	\langle \bm{Y}_n, \bm{a} \rangle = \sum_{i=1}^n \bm{a}_i Y_i \sim N(0,\|\bm{a}\|^2\sigma^2).
	$$
	Suppose $\bm{v}$ is a random vector generated in the way described in 
	Subsection~\ref{subsec:asyminit}.
	Then for any $\bm{x} \in \mathbb{R}^{n+1}$,
	$$
	\langle \bm{v}, \bm{x} \rangle = \sum_{i \ne k} \bm{v}_i \bm{x}_i + \bm{v}_k\bm{x}_k
	= \bm{x}_k \left(Z + \bm{v}_k \right),
	$$
	where $\tilde{\sigma}^2 = \|\tilde{\bm{x}}_{-k}\|^2\sigma^2$
	and $Z \sim N(0,\tilde{\sigma}^2)$.
	If $\bm{x}_k > 0$ and $\|{\x}_{-k}\|^2 > 0$, we have
	$$
	\langle \bm{v}, \bm{x} \rangle > 0 \quad \iff \quad \bm{v}_k > -Z \overset{d}{=} Z.
	$$
	Therefore, it suffices to compute $P(\bm{v}_k > Z)$.
	Let $f_Z(z)$ be the pdf of $Z$ and $f_{\text{P}}(x)$ be the pdf of $\bm{v}_k\sim \text{P}$. Then
	\begin{align*}
	P(\bm{v}_k > Z)
	&= \int_{-\infty}^\infty \int_{z}^M f_{\text{P}}(x)dx f_Z(z)dz \\
	&= \int_{-\infty}^0 \int_{0}^M f_{\text{P}}(x)dx f_Z(z)dz + \int_0^M \int_z^M f_{\text{P}}(x)dx f_Z(z)dz \\
	&= \frac{1}{2} + \int_0^M (1-F_{\text{P}}(z))f_Z(z)dz,
	\end{align*}
	which completes the proof.
\end{proof}

\begin{proof}
For each $\ell$ and $s$, let $k_s^\ell$ be randomly uniformly chosen in $\{1,\cdots,N_{\ell-1}+1\}$.
Let $\bm{V}^{\ell}_s = [\bm{W}^\ell_s, \bm{b}^\ell_s]$
and $\textbf{n}^\ell(\x) = \left[\phi(\N^\ell(\x)),1 \right]$.
Recall that for a vector $\textbf{v} \in \mathbb{R}^{N+1}$,
$$
\textbf{v}_{-k} := [v_1,\cdots,v_{k-1},v_{k+1},\cdots,v_{N+1}].
$$
To emphasize the dependency of $k_s^\ell$,
we denote $\bm{V}^\ell_s$ whose $k$-th component is generated from $\text{P}$
by $\bm{V}^\ell_s(k)$.

The proof can be started from Equation~\ref{app:thm1:eqn4-prob}.
It suffices to compute 
\begin{equation*}
    P(\tilde{A}_{j,\x}|\tilde{A}_{j-1,\x}^c) 
        = \prod_{s=1}^{N_{j-1}} 
        P\left(\bm{W}^{j-1}_s\phi(\N^{j-2}(\x)) + \bm{b}^{j-1}_s \le 0 |\tilde{A}_{j-1,\x}^c\right). 
\end{equation*}
Note that for each $s$,  
$$
P\left(\langle \bm{V}^j_s, \textbf{n}^{j-1} \rangle \le 0 |\tilde{A}_{j-1,\x}^c\right) = 
\frac{1}{N_{j-1}+1} \sum_{k=1}^{N_{j-1}+1}  P\left( \langle \bm{V}^j_s(k), \textbf{n}^{j-1} \rangle \le 0 | \tilde{A}_{j-1,\x}^c\right).
$$
Also, from Lemma~\ref{app:lemma:prob-positive-real}, we have
\begin{align*}
    P\left( \langle \bm{V}^j_s(k), \textbf{n}^{j-1} \rangle \le 0 | \tilde{A}_{j-1,\x}^c\right)
    = \frac{1}{2} - \int_0^{M}
	(1-F_P(z))
	\frac{1}{\sqrt{2\pi}\tilde{\sigma}_{\ell,k}}e^{-\frac{z^2}{2\tilde{\sigma}_{\ell,k}^2}} dz
\end{align*}
where
$$
\tilde{\sigma}_{\ell,k}^2 = \frac{\|\textbf{n}^{\ell-1}_{-k}(\x))\|^2\sigma_\ell^2}{
	\left(\textbf{n}^{\ell-1}_k(\x)\right)^2}.
$$
Note that if $\textbf{n}^{\ell-1}_k(\x) = 0$, the above probability is simply $1/2$.
Also, since the event $\tilde{A}_{j-1,\x}^c$ is given, we know that 
$\phi(\N^\ell(\x)) \ne 0$. Thus, 
$$
P\left( \langle \bm{V}^j_s, \textbf{n}^{j-1} \rangle \le 0 | \tilde{A}_{j-1,\x}^c\right) < \frac{1}{2}.
$$
We thus denote 
$$
\left(\frac{1}{2}-\delta_{\ell-1,\x}(\omega)\right)^{N_{\ell-1}}:=
P(\tilde{A}_{\ell,\x}|\tilde{A}_{\ell-1,\x}^c(\omega)),
$$
where $\delta_{\ell-1,\x}$ is $\mathcal{F}_{\ell-2}$ measurable 
whose value lies in $(0,0.5]$ and it depends on $\x$.
It then follows from Equation~\ref{app:thm1:eqn3} that
\begin{align*}
    P(\tilde{A}_{\ell,\x}^c)
    = \int  \prod_{j=1}^{\ell-1} \left(1 - \left(\frac{1}{2}-\delta_{j,\x}(\omega)\right)^{N_{j}}\right)
    d{P}(\omega)
\end{align*}
where ${P}$ is the probability distribution with respect to $\{\bm{W}^{j}, \bm{b}^{j}\}_{1\le j \le \ell}$.
Note that since the weight matrix and the bias vector of the first layer is initialized from a symmetric distribution, $\delta_{1,\x} = 0$.
Let $\delta_{1,\x}^* = 0$.
By the mean value theorem, there exists some numbers 
$\delta_{2,\x}^*,\cdots, \delta_{\ell-1,\x}^* \in (0,0.5]$ 
such that
$$
P(\tilde{A}_{\ell,\x}^c)
=
\prod_{j=1}^{\ell-1} \left(1- \left(\frac{1}{2}-\delta_{j,\x}^*\right)^{N_{j-1}}\right).
$$
Let $\bm{\delta}^*_{\x} = (\delta_{1,\x}^*,\cdots,\delta_{\ell-1,\x}^*)$.
By setting
$$
\bm{\delta}^* = \bm{\delta}^*_{\x^*}, \qquad \text{where} \quad
\x^*= \argmin_{\x \in \D \backslash \{\bm{0}\}} \hspace{0.1cm} \prod_{j=1}^{\ell-1} \left(1- \left(\frac{1}{2}-\delta_{j,\x}^*\right)^{N_{j-1}}\right),
$$
the proof is completed.
\end{proof}

\section{Proof of Theorem~\ref{thm:2ndmo-asyminit}} \label{app:thm:2ndmo-asyminit}
Let $X_\ell \sim \text{P}_\ell$ whose pdf is denoted by $f_{P_\ell}(x)$.
    Suppose $0 < \mu_{\ell,i}' = E[X_\ell^i] < \infty$ for $i=1,2$.
	We then define three probability distribution functions:
	\begin{equation*}
	f^{(2)}_{P_\ell}(x) = \frac{x^2}{\mu_{\ell,2}'}f_{P_\ell}(x), \qquad
	f^{(1)}_{P_\ell}(x) = \frac{x}{\mu_{\ell,1}'}f_{P_\ell}(x), \qquad
	f^{(0)}_{P_\ell}(x) = f_{P_\ell}(x).
	\end{equation*}
	Also we denote its corresponding cdfs by $F^{(i)}_{P_\ell}(t) = \int_0^t f^{(i)}_{P_\ell}(x)dx$.
	For notational convenience, let us assume that
	$\text{P}_{\ell} = \text{P}$ and $M_\ell = M$ for all $\ell$
	and define
    $$
    \textbf{n}^{\ell}(\x):=\left[\phi(\N^{\ell}(\x)), 1 \right] \in \mathbb{R}_+^{N_{\ell}+1}.
    $$
	We denote the pdf of normal distributions by $f_Y^{\ell,k}(y)$
	where
	\begin{align*}
	f_Y^{\ell,k}(y) = \frac{1}{\sqrt{2\pi}\tilde{\sigma}_{\ell,k}}e^{-\frac{y^2}{2\tilde{\sigma}_{\ell,k}^2}}, \qquad
	\tilde{\sigma}_{\ell,k}^2 = \frac{\|\textbf{n}^{\ell-1}_{-k}(\x))\|^2\sigma_\ell^2}{
	\left(\textbf{n}^{\ell-1}_k(\x)\right)^2},
	\qquad
	\sigma_\ell^2 := \frac{\sigma_w^2}{N_{\ell-1}}
	\end{align*} 
	where $1 \le k \le N_{\ell-1}+1$.
	Recall that for a vector $\textbf{v} \in \mathbb{R}^{N+1}$,
	$$
	\textbf{v}_{-k} := [v_1,\cdots,v_{k-1},v_{k+1},\cdots,v_{N+1}].
	$$
	For any probability distribution $\text{P}$ defined on $[0,M]$ whose pdf is denoted by $f_P(x)$, let
    \begin{equation} \label{app:thm:2ndmo-asyminit-defJ}
        \mathcal{J}^{\ell,k}_{\text{P}}:= \int_0^M \int_y^M f_P(x)dx f_Y^{\ell,k}(y)dy. 
    \end{equation}
    Let $Y_{\ell,k} \sim N(0,\tilde{\sigma}_{\ell,k}^2)$
    and $X \sim \text{P}$.
    Then 
    \begin{align*}
        P\left(X > Y_{\ell,k} | \tilde{\sigma}_{\ell,k}^2\right)
    &= \int_{-\infty}^\infty \int_{0}^M \mathds{1}_{\{X >Y_{\ell,k}\}} f_{P}(x) f^{\ell,k}_Y(y) dxdy \\
    &= \int_{-\infty}^0 \int_{0}^M f_P(x)dx f^{\ell,k}_Y(y)dy
    +
    \int_{0}^M \int_{y}^M f_P(x) dxf^{\ell,k}_Y(y)dy \\
    &= \frac{1}{2} +\mathcal{J}^{\ell,k}_{\text{P}}.
    \end{align*}
    Therefore, $0 \le \mathcal{J}^{\ell,k}_{\text{P}} \le 0.5$.
	
The proof of Theorem~\ref{thm:2ndmo-asyminit} starts with the following lemma.
\begin{lemma} \label{app:lemma-J}
	Suppose $Y \sim N(0,\tilde{\sigma}_{\ell,k}^2)$,
	$X \sim \text{P}_\ell$ and $\mu_{\ell,1}' \ge M/2$.
    Then
	\begin{equation*}
	\begin{split}
	\int_{0}^M \int_{y}^M (x-y)^2 f_{P_\ell}(x)dx f_Y^{\ell,k}(y)dy 
	\le \mu_{\ell,2}'/2.
	\end{split}
	\end{equation*}
\end{lemma}
\begin{proof}
	Let $I_{\ell,k}$ be the quantity of our interest:
	\begin{align*}
	I_{\ell,k}:=\int_{0}^M \int_{y}^M (x-y)^2 f_{P_\ell}(x)dx f_Y^{\ell,k}(y)dy.
	\end{align*}
	Without loss of generality, we can assume $M=1$.
    This is because 
    \begin{align*}
        I_{\ell,k} &= \int_0^M \int_y^M (x-y)^2 f_{P_\ell}(x) f_Y^{\ell,k}(y) dxdy
        \\
        &= \int_0^M \int_{y/M}^1 (Mu-y)^2 f_{P_\ell}(Mu) f_Y^{\ell,k}(y) Mdu dy \\
        &= \int_0^1 \int_{s}^1 M^2(u-s)^2 (Mf_{P_\ell}(Mu)) (Mf_Y^{\ell,k}(Ms)) du ds \\
        &= M^2 \int_0^1 \int_s^1 (u-s)^2 f_{U}(u) f_{S}^{\ell,k}(s) du ds \\
        &= M^2 \tilde{I}_{\ell,k}
    \end{align*}
    where $u = x/M$ and $s = y/M$. 
    Here $MU \sim \text{P}_\ell$ and 
    $S \sim N(0,\tilde{\sigma}^2_{\ell,k}/M^2)$.
    
	Let us first consider the inner integral.
	Let $G(y)=\int_{y}^1 (x^2-2xy+y^2) f_{P_\ell}(x) dx$.
	It then can be written as follow:
	\begin{align*}
	G(y)&= \int_{y}^1 (x^2-2xy+y^2) f_{P_\ell}(x) dx 
	= \int_{y}^1 (x^2f_{P_\ell}(x) -2yxf_{P_\ell}(x)+y^2f_{P_\ell}(x)) dx\\
	&= \int_y^1 \left(\mu_{\ell,2}' f^{(2)}_{P_\ell}(x) - 2\mu_{\ell,1}'y f^{(1)}_{P_\ell}(x) + y^2f^{(0)}_{P_\ell}(x) \right) dx
	\\
	&= \mu_{\ell,2}'(1 - F^{(2)}_{P_\ell}(y)) -2y\mu_{\ell,1}'(1 - F^{(1)}_{P_\ell}(y)) + y^2(1-F^{(0)}_{P_\ell}(y)) \\
	&= -2\mu_{\ell,1}'y + y^2
	+ \mu_{\ell,2}'\int_y^1 f^{(2)}_{P_{\ell}}(x)dx 
	+2\mu_{\ell,1}' yF^{(1)}_{P_\ell}(y) - y^2F^{(0)}_{P_\ell}(y).
	\end{align*}
	Note that since $0 \le y \le 1$, and thus $y^2 \le y$,
	we have
	\begin{align*}
	    G(y) &= y^2(1-F^{(0)}_{P_\ell}(y)) - 2\mu_{\ell,1}'y(1-F^{(1)}_{P_\ell}(y)) + \mu_{\ell,2}'\int_y^1 f^{(2)}_{P_{\ell}}(x)dx\\
	    &\le y(1-F^{(0)}_{P_\ell}(y)) - 2\mu_{\ell,1}'y(1-F^{(1)}_{P_\ell}(y)) + \mu_{\ell,2}'\int_y^1 f^{(2)}_{P_{\ell}}(x)dx \\
	    &= -y(2\mu_{\ell,1}'-1)\left[1 - \left(\frac{2\mu_{\ell,1}'F^{(1)}_{P_\ell}(y) - F^{(0)}_{P_\ell}(y)}{2\mu_{\ell,1}'-1}\right)\right]
	    + \mu_{\ell,2}'\int_y^1 f^{(2)}_{P_{\ell}}(x)dx.
	\end{align*}
	Since $2\mu_{\ell,1}' \ge 1$ and
	$$
	1 - \left(\frac{2\mu_{\ell,1}'F^{(1)}_{P_\ell}(y) - F^{(0)}_{P_\ell}(y)}{2\mu_{\ell,1}'-1}\right) \ge 0, \qquad \forall y \in [0,1],
	$$
	we have $G(y) \le \mu_{\ell,2}'\int_y^1 f^{(2)}_{P_{\ell}}(x)dx$.
	Therefore,
	\begin{align*}
	    I_{\ell,k} \le \mu_{\ell,2}' \int_0^1 \int_y^1 f^{(2)}_{P_{\ell}}(x)dx f^{\ell,k}_Y(y)dy= \mu_{\ell,2}' \mathcal{J}^{\ell,k}_{f^{(2)}_{P_\ell}}
	\end{align*}
	where $\mathcal{J}^{\ell,k}_{\text{P}}$ is defined in Equation~\ref{app:thm:2ndmo-asyminit-defJ}.
	Since $0\le \mathcal{J}^{\ell,k}_{\text{P}} \le 0.5$,
	we conlcude that
	$I_{\ell,k} \le \mu_{\ell,2}'/2$.
\end{proof}

\begin{proof}
	It follows from
	$$
	\N^{\ell}_j(\x) = \bm{W}_j^{\ell}\phi(\N^{\ell-1}(\x)) + \bm{b}^{\ell}_j, \qquad 1\le j \le N_\ell
	$$
	that we have
	\begin{align*}
	E\left[\N^{\ell}_j(\x)^2 | \N^{\ell-1}(\x) \right]
	= \sigma^2_\ell \sum_{t \ne k_j^\ell} \phi(\N^{\ell-1}_t(\x))^2
	+ \mu_{\ell,2}'\phi(\N^{\ell-1}_{k_j^\ell}(\x))^2.
	\end{align*}
	Since $k_j^\ell$ is randomly uniformly selected, 
	by taking the expectation with respect to $k_j^\ell$, we have
	\begin{align*}
	\mathbb{E}\left[\N^{\ell}_j(\x)^2 | \N^{\ell-1}(\x) \right]
	=\frac{\sigma_\ell^2N_{\ell} + \mu_{\ell,2}'}{N_{\ell}+1}\left(1+
		\|\phi(\N^{\ell-1}(\x))\|^2 \right).
	\end{align*}
	Since the above is independent of $j$, 
	and $q^\ell(\x) = \|\N^\ell(\x)\|^2/N_\ell$,
	\begin{equation} \label{app:length-map-main-eqn}
	\mathbb{E} \left[ q^{\ell}(\x) | \N^{\ell-1}(\x) \right]
	= \frac{\sigma_\ell^2N_{\ell} +\mu_{\ell,2}'}{N_{\ell}+1}\left(1+
	\|\phi(\N^{\ell-1}(\x))\|^2 \right).
	\end{equation}
	At a fixed $\ell$ and for each $t=1,\cdots, N_\ell$,
	let $\textbf{v}^{\ell,t}=[\textbf{W}^\ell_t, \textbf{b}^\ell_t] \in \mathbb{R}^{N_{\ell-1}+1}$ be 
	a random vector such that 
	$\textbf{v}_{-k_t}^{\ell,t} \sim N(0,\sigma_\ell^2 \bm{I}_{N_{\ell-1}})$
	and $\textbf{v}_{k_t}^{\ell,t} \sim \text{P}_\ell$.
	Then $\N^\ell_t(\x) = \langle \textbf{v}^{\ell,t}, \textbf{n}^\ell(\x) \rangle$.
    Given $\N^{\ell-1}(\x)$, assuming $(\textbf{n}^{\ell-1}_{k_t}(\x)) > 0$, one can view $\N^{\ell}_t(\x)$ as
    $$
    \N^\ell_t(\x) = \textbf{n}^{\ell-1}_{k_t}(\x)\left(\sigma_{\ell,k_t}Z + X_\ell\right), \qquad
    Z \sim N(0,1) \text{ and } X_\ell \sim \text{P}_\ell.
    $$
    Thus $\phi(\N^\ell_t(\x))^2 = \left(\textbf{n}^{\ell-1}_{k_t}(\x)\right)^2\phi\left(\sigma_{\ell,k_t}Z + X_\ell\right)^2$
    and we obtain 
	\begin{align*}
	&\frac{1}{(\textbf{n}^{\ell-1}_{k_t}(\x))^2}E\left[\phi(\N^{\ell}_t(\x))^2 | \N^{\ell-1}(\x) \right]
	=E\left[\phi\left(\sigma_{\ell,k_t}Z + X_\ell\right)^2 | \N^{\ell-1}(\x) \right]
	\\
	&= \int_{-\infty}^M \int_{y}^M (x-y)^2 dF_P(x) f_Y^{\ell,k_t}(y)dy \\
	&= \int_{-\infty}^0 \int_{0}^M (x-y)^2 dF_P(x) f_Y^{\ell,k_t}(y)dy
	+ \int_{0}^M \int_{y}^M (x-y)^2 dF_P(x) f_Y^{\ell,k_t}(y)dy \\
	&= \int_{-\infty}^0 \int_{0}^M \left(x^2+y^2-2xy\right) dF_P(x) f_Y^{\ell,k_t}(y)dy 
	+ \int_{0}^M \int_{y}^M (x-y)^2 dF_P(x) f_Y^{\ell,k_t}(y)dy \\
	&= \int_{-\infty}^0 \left(\mu_{\ell,2}' + y^2 - 2\mu_{\ell,1}'y\right) f_Y^{\ell,k_j}(y)dy
	+ \int_{0}^M \int_{y}^M (x-y)^2 dF_P(x) f_Y^{\ell,k_t}(y)dy \\
	&= \frac{1}{2}\left(\mu_{\ell,2}' + \tilde{\sigma}_{\ell,k_t}^2 \right)
	+ \mu_{\ell,1}'\sqrt{\frac{2}{\pi}}\tilde{\sigma}_{\ell,k_t}
	+ \int_{0}^M \int_{y}^M (x-y)^2 dF_P(x) f_Y^{\ell,k_t}(y)dy.
	\end{align*}
	By multiplying $(\textbf{n}^{\ell-1}_{k_t}(\x))^2$ in the both sides, we have 
	\begin{align*}
	E[\phi(\N^{\ell}_t(\x))^2 | \N^{\ell-1}(\x) ] &= 
	\frac{1}{2}\left((\mu_{\ell,2}'-\sigma_{\ell}^2)(\textbf{n}^{\ell-1}_{k_t}(\x))^2 + (\|\phi(\N^{\ell-1}(\x))\|^2 + 1)\sigma_{\ell}^2 \right)
	\\
	&\qquad
	+ (\textbf{n}^{\ell-1}_{k_t}(\x))^2\int_{0}^M \int_{y}^M (x-y)^2 dF_P(x) f_Y^{\ell,k_t}(y)dy \\
	&\qquad\qquad 
	+ \mu_{\ell,1}'\sigma_\ell \sqrt{\frac{2}{\pi}} \textbf{n}^{\ell-1}_{k_t}(\x)
	\|\textbf{n}^{\ell-1}_{-k_t}(\x)\|.
	\end{align*}
	Since $k_t$ is randomly selected, by taking expectation w.r.t. $k_j$, we have
	\begin{equation} \label{app:thm:2ndmo-asyminit-eqn1}
	\begin{split}
	    \mathbb{E}[\phi(\N^{\ell}_t(\x))^2 | \N^{\ell-1}(\x) ] &= 
	\frac{(1 + \|\phi(\N^{\ell-1}(\x))\|^2)}{2}\left(\frac{\mu_{\ell,2}'-\sigma_\ell^2}{N_{\ell-1}+1} + \sigma_\ell^2\right) \\
	&\qquad + \sum_{k_t=1}^{N_{\ell-1}+1}\frac{(\textbf{n}^{\ell-1}_{k_t}(\x))^2}{N_{\ell-1}+1}\int_{0}^M \int_{y}^M (x-y)^2 dF_P(x) f_Y^{\ell,k_t}(y)dy
	\\
	&\qquad\qquad 
	+ \mu_{\ell,1}'\sqrt{\frac{2}{\pi}}\sigma_\ell 
	\sum_{k_t=1}^{N_{\ell-1}+1} \frac{\textbf{n}^{\ell-1}_{k_t}(\x)
	\|\textbf{n}^{\ell-1}_{-k_t}(\x)\|}{N_{\ell-1}+1}.
	\end{split}
	\end{equation}
	By the Cauchy-Schwarz inequality, the third term in the right hand side of Equation~\ref{app:thm:2ndmo-asyminit-eqn1} can be bounded by
	\begin{align*}
	\mu_{\ell,1}'\sqrt{\frac{2}{\pi}}\sigma_\ell \sum_{k_t=1}^{N_{\ell-1}+1} \frac{\textbf{n}^{\ell-1}_{k_t}(\x)
	\|\textbf{n}^{\ell-1}_{-k_t}(\x)\|}{N_{\ell-1}+1}
	\le \mu_{\ell,1}'\sqrt{\frac{2}{\pi}}\frac{\sigma_w(1 + \|\phi(\N^{\ell-1}(\x))\|^2)}{N_{\ell-1}+1},
	\end{align*}
	where $\sigma_w = \sigma_\ell \sqrt{N_{\ell-1}}$.
	Let 
	$I_{\ell,k_t} = \int_{0}^M \int_{y}^M (x-y)^2 dF_P(x) f_Y^{\ell,k_t}(y)dy$.
	It then follows from Lemma~\ref{app:lemma-J}
	that 
	$I_{\ell,k} \le \mu_{\ell,2}'/2$ for any $k$.
	Thus the second term in the right hand side of Equation~\ref{app:thm:2ndmo-asyminit-eqn1} can be bounded by
	\begin{align*}
	    \sum_{k_t=1}^{N_{\ell-1}+1}\frac{(\textbf{n}^{\ell-1}_{k_t}(\x))^2}{N_{\ell-1}+1}\int_{0}^M \int_{y}^M (x-y)^2 dF_P(x) f_Y^{\ell,k_t}(y)dy
	    \le 
	    \frac{\mu_{\ell,2}'}{2} \frac{\left(1+\|\phi(\N^{\ell-1}(\x))\|^2\right)}{N_{\ell-1}+1}
	\end{align*}
	Therefore, we obtain
	\begin{align*}
	E[\phi(\N^{\ell}_t(\x))^2 | \N^{\ell-1}(\x) ] 
	\le 
	\mathcal{C}_\ell \frac{(1 + \|\phi(\N^{\ell-1}(\x))\|^2)}{2}
	\end{align*}
	where
	$$
	\mathcal{C}_\ell = \frac{\mu_{\ell,2}'-\sigma_\ell^2}{N_{\ell-1}+1} + \sigma_\ell^2 +
	\frac{2\sqrt{2}\mu_{\ell,1}'\sigma_w}{(N_{\ell-1}+1)\sqrt{\pi}}
	+\frac{\mu_{\ell,2}'}{N_{\ell-1}+1}.
	$$
	Since $\mathcal{C}_\ell$ is independent of $t$,
	$$E[\|\phi(\N^{\ell}(\x))\|^2 | \N^{\ell-1}(\x) ]
	\le N_\ell \mathcal{C}_\ell \frac{(1 + \|\phi(\N^{\ell-1}(\x))\|^2)}{2}.
	$$
	It then follows from Equation~\ref{app:length-map-main-eqn},
	\begin{align*}
	1+	\|\phi(\N^{\ell-1}(\x))\|^2
	=\frac{N_{\ell}+1}{\sigma_\ell^2N_{\ell} + \mu_{\ell,2}'} E\left[q^{\ell}(\x) | \N^{\ell-1}(\x) \right]
	\end{align*}
	that
	\begin{align*}
	E[\|\phi(\N^{\ell}(\x))\|^2 | \N^{\ell-1}(\x) ]
	\le 
	\frac{\mathcal{C}N_{\ell}(N_{\ell}+1)}{2(\sigma_\ell^2N_{\ell}+ \mu_{\ell,2}')} E\left[q^{\ell}(\x) | \N^{\ell-1}(\x) \right].
	\end{align*}
	Thus we have
	\begin{align*}
	E\left[q^{\ell+1}(\x) | \N^{\ell-1}(\x) \right] 
	&= \frac{\sigma_{\ell+1}^2N_{\ell+1} + \mu_{\ell+1,2}'}{N_{\ell+1}+1}\left(1+
	E[\|\phi(\N^{\ell}(\x))\|^2 | \N^{\ell-1}(\x) ] \right) \\
	&\le 
	\sigma_{b,\ell}^2
	+ \frac{\mathcal{A}_{\ell,upp}}{2}
	E\left[q^{\ell}(\x) | \N^{\ell-1}(\x) \right]
	\end{align*}
	where
	\begin{align*}
	    \sigma_{b,\ell}^2 = \frac{\sigma_{\ell+1}^2N_{\ell+1} + \mu_{\ell+1,2}'}{N_{\ell+1}+1}, \qquad
	    \mathcal{A}_{\ell,upp} = \frac{\sigma_{b,\ell}^2 N_{\ell}(N_{\ell}+1)}{\sigma_\ell^2N_{\ell} + \mu_{\ell,2}'}\mathcal{C}_\ell.
	\end{align*}
	By taking expectation with respect to $\N^{\ell-1}(\x)$,
	we obtain
	\begin{align*}
	E[q^{\ell+1}(\x)]&\le 
	\frac{\mathcal{A}_{\ell,upp}}{2} E[q^{\ell}(\x)]
	+ \sigma_{b,\ell}^2
	\end{align*}
	The lower bound can be obtained by dropping the second and the third terms in Equation~\ref{app:thm:2ndmo-asyminit-eqn1}.
	Thus
	\begin{align*}
	\frac{\mathcal{A}_{\ell,low}}{2}
	E[q^{\ell}(\x)]
	+ \sigma_{b,\ell}^2
	\le
	E[q^{\ell+1}(\x)]
	\le 
	\frac{\mathcal{A}_{\ell,upp}}{2} E[q^{\ell}(\x)]
	+ \sigma_{b,\ell}^2
	\end{align*}
	where
	$$
	\mathcal{A}_{\ell,low} =
	\frac{\sigma_{\ell+1}^2N_{\ell+1} + \mu_{\ell+1,2}'}{N_{\ell+1}+1}
	\frac{N_{\ell}+1}{\sigma_\ell^2N_{\ell} + \mu_{\ell,2}'} 
	\left(\frac{\mu_{\ell,2}'N_\ell-\sigma_w^2}{N_{\ell-1}+1} + \sigma_w^2\right).
	$$
\end{proof}

\section{Randomized asymmetric initialization} \label{app:RAI_code}

\begin{lstlisting}
import numpy as np


def RAI(fan_in, fan_out):
    """Randomized asymmetric initializer.
    It draws samples using RAI where fan_in is the number of input units in the weight tensor and fan_out is the number of output units in the weight tensor.
    """
    V = np.random.randn(fan_out, fan_in + 1) * 0.6007 / fan_in ** 0.5
    for j in range(fan_out):
        k = np.random.randint(0, high=fan_in + 1)
        V[j, k] = np.random.beta(2, 1)
    W = V[:, :-1].T
    b = V[:, -1]
    return W.astype(np.float32), b.astype(np.float32)
\end{lstlisting}

\bibliographystyle{plain}
\bibliography{main}

\end{document}